\documentclass[11pt,a4paper]{article}
\usepackage[hyperref]{emnlp2020}
\usepackage{times}
\usepackage{latexsym}
\usepackage{enumitem}

\setitemize{noitemsep,topsep=0pt,parsep=0pt,partopsep=0pt}
\setenumerate{noitemsep,topsep=0pt,parsep=0pt,partopsep=0pt}

\usepackage{microtype}

\aclfinalcopy % Uncomment this line for the final submission
\usepackage{tkz-kiviat} 
\usetikzlibrary{arrows}

\usepackage{booktabs, tabularx}
\usepackage{amssymb}
\usepackage{subfig}

\usepackage{pgfplots}
\usepgfplotslibrary{fillbetween}
\usetikzlibrary{patterns}
\usetikzlibrary{pgfplots.groupplots}
\pgfplotsset{compat=1.17}
\usepackage{lscape}
\usepackage{multirow}
\usepackage{dblfloatfix}

\newcommand\blfootnote[1]{%
  \begingroup
  \renewcommand\thefootnote{}\footnote{#1}%
  \addtocounter{footnote}{-1}%
  \endgroup
}

\usepackage{hyphenat}
\DeclareCaptionTextFormat{new}{\nohyphens{#1}}
\title{Not All Models Localize Linguistic Knowledge in the Same Place:\\ A Layer-wise Probing on BERToids' Representations}

\author{
    Mohsen Fayyaz$^{1\star}$ ~ Ehsan Aghazadeh$^{1\star}$ ~ Ali Modarressi$^{2}$ ~ Hosein Mohebbi$^{2}$\\ ~ \textbf{Mohammad Taher Pilehvar$^{3}$} \\
    $^1$ University of Tehran, Iran ~
    $^2$ Iran University of Science and Technology, Iran \\
    $^3$ Tehran Institute for Advanced Studies, Khatam University, Iran \\
    \texttt{\{mohsen.fayyaz77, eaghazade1998\}@ut.ac.ir}\\
    \texttt{\{m\_modarressi, hosein\_mohebbi\}@comp.iust.ac.ir}\\
    \texttt{mp792@cam.ac.uk}
  }

\date{}

\begin{document}
\maketitle
\begin{abstract}

% Background
Most of the recent works on probing representations have focused on BERT, with the presumption that the findings might be similar to the other models.
% Purpose
In this work, we extend the probing studies to two other models in the family, namely ELECTRA and \nohyphens{XLNet}, showing that variations in the pre-training objectives or architectural choices can result in different behaviors in encoding linguistic information in the representations.
Most notably, we observe that ELECTRA tends to encode linguistic knowledge in the deeper layers, whereas \nohyphens{XLNet} instead concentrates that in the earlier layers.
Also, the former model undergoes a slight change during fine-tuning, whereas the latter experiences significant adjustments.
Moreover, we show that drawing conclusions based on the \emph{weight mixing} evaluation strategy---which is widely used in the context of layer-wise probing---can be misleading given the norm disparity of the representations across different layers.
Instead, we adopt an alternative information-theoretic probing with \emph{minimum description length}, which has recently been proven to provide more reliable and informative results.

\blfootnote{$^\star$Equal contribution.}
\end{abstract}

% INTRO ----------------------------
\section{Introduction}
With the impressive success of pre-trained language models, such as BERT \citep{devlin-etal-2019-bert}, and their significant advances in transfer learning, a wave of interest has recently been directed toward understanding the knowledge encoded in their representations \citep{rogers-etal-2020-primer}.
One of the analytical tools which is widely used for this investigation is \emph{probing}: training a shallow supervised classifier that attempts to predict specific linguistic properties or reasoning abilities, based on representations obtained from the model \citep{tenney2018you, tenney-etal-2019-bert, hewitt-manning-2019-structural, talmor2020olmpics, mohebbi2021exploring, ushio-etal-2021-bert, chen2021probing}.

However, most of the previous studies have focused on BERT only, neglecting other models in the family.
This leaves open the question of how training objectives (which are fundamentally different for some models) and architectural choices would impact the resulting representations and the knowledge encoded in them.

In this work, we carry out an analysis on three popular language models with totally different pre-training objectives: BERT (masked language modeling), \nohyphens{XLNet} (permuted language modeling, \citealp{yang2019xlnet}), and ELECTRA (replaced token detection, \citealp{Clark2020ELECTRA}).
We also show that the ``weight mixing'' evaluation strategy of \citet{tenney-etal-2019-bert}, which is widely used in the context of probing \citep[\textit{inter alia}]{de-vries-etal-2020-whats, kuznetsov-gurevych-2020-matter, choenni2020does}, might not be a reliable basis for drawing conclusions in the layer-wise cross model analysis as it does not take into account the norm disparity across the representations of different layers.
Instead, we perform an information-theoretic probing analysis using Minimum Description Length proposed by \citet{voita-titov-2020-information}.

Based on a series of experiments, we find that language models derived from BERT have different behaviors in encoding linguistic knowledge. 
Specifically, we show that, unlike BERT, \nohyphens{XLNet} encodes linguistic information in the earlier layers during pre-training, while ELECTRA tends to carry this information to the higher layers.
We also extend our probing experiments to the fine-tuned setting to assess the extent of change in the encoded knowledge upon fine-tuning.
Using Representation Similarity Analysis \cite[RSA]{neuroscience-rsa}, we show that representations from the higher layers in \nohyphens{XLNet}---that do not well encode specific linguistic knowledge---undergo substantially bigger changes during fine-tuning when compared to the other models. 
In summary, our main contributions are as follows:
\begin{itemize}
    \item We point out that the weight mixing evaluation strategy in edge probing does not lead to reliable conclusions in layer-wise cross model analysis studies.
    \item By relying on an information-theoretic probing method, we carry out a probing analysis across three commonly used pre-trained models.
    \item We also extend our probing experiments to fine-tuned representations to examine how linguistic information changes during fine-tuning.
    \item To provide complementary results to validate our findings, we also employ RSA to measure the amount of change in the representations after fine-tuning.
\end{itemize}

% BACKGROUND -------------------------------------------
\section{Background and Pilot Analysis}
In this section, we review BERT and two of its popular derivatives: XLNet and ELECTRA, highlighting their differences as well as two commonly used probing methods that are essential to our discussion.
We then conduct a pilot experiment to show the limitations of the evaluation metric used in edge probing and then justify our choice of an information-theoretic alternative.

\subsection{Models}
There is a wide variety of models derived from BERT, which are generally categorized into \emph{autoencoding} and \emph{autoregressive} models. Here, we focus on the most prominent model in each category. The two models have totally different pre-training objectives and have both shown outstanding performance on standard NLP tasks. 
BERT\footnote{Due to resource limitations, our evaluations are based on the \textit{base} version (12-layer, 768-hidden size, 12-attention head, 110M parameters) of each model obtained from the HuggingFace's Transformers library \citep{wolf-etal-2020-transformers}.} is also considered as our baseline.
\paragraph{BERT.}
BERT \cite{devlin-etal-2019-bert} has multiple Transformer encoder layers \citep{vaswani2017attention} stacked on top of each other, which are pre-trained with two self-supervised training objectives; Masked Language Model (MLM) and Next Sentence Prediction (NSP). The former predicts randomly masked tokens in the input sentence, whereas the latter checks whether two sentences could be considered consecutive.

\paragraph{XLNet.}
In contrast to BERT, which attempts to reconstruct the original sentence from corrupted input, \nohyphens{XLNet} is an auto-regressive model based on the Transformer-XL architecture \cite{dai-etal-2019-transformer} that leverages permutation language modeling to learn from a bi-directional context. 
This allows the model to consider dependencies between masked tokens in a sentence for prediction.
Due to the permuted order of the tokens in this objective, the next predicted token could occur at any position, making it a more difficult task. 
To address this, \nohyphens{XLNet} uses a two-stream self-attention mechanism (one for content and one for the query) to autoregressively predict tokens.
\nohyphens{XLNet} has also excluded NSP during pre-training. 
Though these modifications resulted in better performance than BERT, the costs of pre-training have increased in terms of FLOPs \cite{Clark2020ELECTRA}.

\paragraph{ELECTRA.}
Another model based on BERT is ELECTRA which falls into the category of autoencoder models. \citet{Clark2020ELECTRA} have introduced the replaced token detection pre-training objective to substitute BERT's objectives. 
ELECTRA jointly trains two models: generator and discriminator.
The generator receives corrupted inputs with masked tokens in the input and tries to reconstruct the input like BERT. The discriminator is trained to predict whether each token was replaced by the generator or not.
The authors pointed out that this novel pre-training objective could lead to learning representations that outperform prior models while requiring less computational cost during training.

% \subsection{Probes}
% \section{Methodology}

\subsection{Edge Probing}
\citet{tenney2018you} introduced edge probing as a means to measure linguistic knowledge in word representations. This is done by training a simple classifier on top of the representations. The accuracy of this classifier is taken as a representative for the quality of the encoded information about the specific task.
Edge probing consists of a set of span-level tasks, where the span is a part of the sentence indicated by the dataset. 
The probe can only access the representations in that specific span. 
Eight labeling tasks are considered, including syntactic tasks, such as dependency labeling, and semantic tasks, such as coreference resolution. 
Before giving the inputs to the classifier, edge probing structure pools the representations across layers to make a fixed size vector for feeding to the classifier.
The work of \citet{tenney-etal-2019-bert} is one of the first studies which leverages edge probing to quantify where linguistic knowledge is captured within BERT. 

\paragraph{Scalar Mixing Weights.}
To estimate the contribution of each layer to a given probing task, \citet{tenney-etal-2019-bert} used a technique called scalar mixing weights \citep{peters-etal-2018-deep} which associates a trainable scalar weight with each layer in the model.
After learning these weights alongside the probing classifier, they interpret layers with higher weights as those having more information for the particular task.

\subsubsection{Mixing Weights Reliability Issues}
\label{se:sanity_edge}
While \citet{tenney-etal-2019-bert} made interesting conclusions on BERT using edge probing and the \emph{scalar mixing weights} evaluation strategy, we argue that this procedure is not reliable for layer-wise comparison. 
Several recent studies have conducted their experiments based on edge probing and \emph{weight mixing} evaluation strategy.
In one such study, \citet{toshniwal-etal-2020-cross} concluded that \nohyphens{XLNet} relies heavily on the input embedding layer in mixing weight evaluation for the coreference arc prediction task \cite{toshniwal-etal-2020-cross}.
We show that this conclusion might not be accurate given that the representation norms in \nohyphens{XLNet} drastically change throughout layers.

\begin{figure}[t]
\centering
    \begin{tikzpicture}
    \begin{axis}[
        width=0.44\textwidth,
        xlabel={Layer},
        ylabel={Representations Norm},
        title style={font=\small},
        label style={font=\small},
        legend style={font=\small},
        % xmin=0, %xmax=100,
        ymin=-10, %ymax=120,
        xtick={0,2,4,6,8,10,12},
        legend pos=north west,
        grid=both,
    ]
    
    \addplot[
        color=blue,
        mark=square*,
        thick
        ]
        coordinates {
        (0, 15.503698)(1, 18.51175267)(2, 20.33464733)(3, 19.92714233)(4, 20.68578233)(5, 21.38014033)(6, 21.76447067)(7, 21.43070267)(8, 20.18465167)(9, 19.17934167)(10, 20.26221567)(11, 20.58813833)(12, 14.86458533)
        };
        
    \addplot[
        color=magenta,
        mark=*,
        thick,
        ]
        coordinates {
        (0, 1.184295233)(1, 24.67719033)(2, 23.14850633)(3, 27.53560333)(4, 28.061238)(5, 22.13762467)(6, 33.75330433)(7, 33.276676)(8, 52.554419)(9, 47.530539)(10, 66.50901467)(11, 122.1762867)(12, 185.3667533)
        };
    
    \legend{BERT, XLNET}
        
    \end{axis}
    \end{tikzpicture}
    
    \caption{Comparison of the representations norm in different layers of XLNet and BERT when tested on Wikipedia examples.
    % In contrast to BERT, 
    XLNet shows considerable norm disparities across different layers.
    }
    \label{fig:layers_norm}
\end{figure}
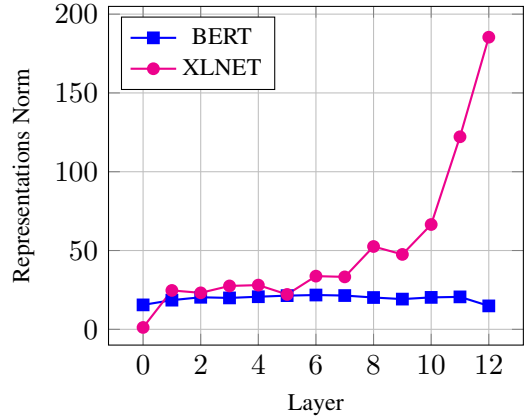

Specifically, Figure~\ref{fig:layers_norm} shows the representation norms across different layers in BERT and \nohyphens{XLNet}. 
These norms are computed based on the representations of 500 tokens sampled from the OPUS dataset \cite{tiedemann-2012-parallel}.
The results are the average of three runs, and the same tokens are given to both models in each run\footnote{If a word was split into more than one token by the tokenizer, we used the first token of that word.}.
In \nohyphens{XLNet}, the norm of the embedding layer is extremely smaller than that of other layers. 
This clearly shows that the concentration of edge probing's weight on the embedding layer does not indicate the level of information encoded in that layer.
Rather, the model tries to compensate for relatively small representation norms.
On the contrary, BERT retains the same level of representation norms across different layers. 
However, even such minor differences in representation norms might affect the conclusions of edge probing.\footnote{To mitigate this issue, one can normalize the representations just before applying the scalar mixing weights to obtain more reliable results, especially in \nohyphens{XLNet}. We reported results with this modification in the appendix (Figure \ref{fig:edge_probing_layers}).}
Given this issue with edge probing, we opted for a theoretically justified method, Minimum Description Length (MDL) probing, in our layer-wise analysis. 

\begin{table*}[h!t!]
\centering
\begin{tabular}{ll}
\toprule
Dependencies & 
I think it will \textbf{[help]$_2$} \textbf{[me]$_1$} very much in my role . $\rightarrow$ obj (object)
\\
NER & 
thirty eight years ago founded \textbf{[the special Olympics]} . $\rightarrow$ EVENT
\\
SRL & 
Their father \textbf{[called]$_1$} later \textbf{[to see if they were fine]$_2$} .
$\rightarrow$ ARGM-PRP (Purpose) 
\\
Coreference &  
% \textbf{[The stolen Enigma machine returned by mail to the BBC]$_2$} is missing three of \textbf{[its]$_1$} coding wheel .
Thank \textbf{[you]$_1$} very much , \textbf{[Tony]$_2$} .
$\rightarrow$ True
\\
Rel. (SemEval) & 
NASA Kepler mission sends \textbf{[names]$_1$} into \textbf{[space]$_2$} .
$\rightarrow$ Entity-Destination(e$_1$,e$_2$)\\
\bottomrule
\end{tabular}
\caption{Examples of sentences, spans, and target labels for each probing task.}
\label{tab:datasets_examples}
\end{table*}
\subsection{MDL Probing}
\label{se:background_mdl}
Conventional probes (such as \citealp{conneau-etal-2018-cram, tenney2018you, jawahar2019does}) leave unclear whether the classifier identifies linguistic knowledge in the representations or learns the task itself \citep{hewitt-liang-2019-designing}.
Hence, researchers had to limit the size of the dataset \cite{zhang-bowman-2018-language} or the probe's complexity \cite{liu-etal-2019-linguistic} to make sure the probe is not learning the task itself. However, probes based on information theory enable us to obtain more interpretable and reliable probing results.

The goal of information-theoretic probing is to measure to what extent representations encode a specific linguistic knowledge and how much effort is required to extract it.
\citet{voita-titov-2020-information} combined the final quality of the probe classifier and the difficulty of achieving it by reformulating probes to a data transmission problem.
If $N$ number of representations are given, we plan to send their corresponding labels with a minimum description length, where each label has $K$ classes.
In \emph{uniform encoding}, we assume that each representation has a label with a probability of $1/K$ and transmit them as raw information without training, which results in the maximum codelength possible of $N\cdot\log_2(K)$.
But suppose the representations show some degree of regularity with respect to the labels\footnote{Highly regulated data would result in shorter codelength than weakly regulated ones \citep{voita-titov-2020-information}.}. In that case, instead of sending the labels, we can train a classifier to predict the labels given the representations and transmit the classifier's complexity (classifier codelength).
Given that the classifier is usually not optimal, the final cross-entropy of the classifier over the data (data codelength) will be added to the classifier's codelength, resulting in an alternative evaluation metric known as MDL.
Also, since the number of targets $N$ will affect the total sum of cross-entropy, and in turn the final codelength (\nohyphens{MDL}), it is preferred to use the compression evaluation metric, which is defined as:
\begin{equation}
\mathbf{c} = \frac{N\cdot\log_2(K)}{\textsc{MDL}}
\label{eq:compression}
\end{equation}
where the uniform codelength is divided by the MDL to eliminate the effect of $N$.
Lower MDL indicates that the classifier predicts labels more accurately than a random guessing classifier, leading to higher compression. In contrast, when our classifier makes a random decision, MDL will be equal to the term in the numerator, resulting in no compression ($\mathbf{c} = 1$).
We will report compression instead of codelength in our experimental results.

To compute MDL, \citet{voita-titov-2020-information} proposed two methods. The first one is \emph{variational coding}, which estimates the complexity of the probe using a Bayesian model. The second method is called \emph{online coding}, which is based on the fact that if the regularity in data is strong, it can be revealed by only a small portion of the data. In this method, the probe will be progressively trained on different amounts of data from small to large portions, and the cross-entropy of each portion will be added together to form the final codelength. 
Since \citet{voita-titov-2020-information} showed that the two compression methods conform in results, we employ the \emph{online coding} method due to its more straightforward implementation.

As the MDL probe is more stable and informative than other conventional probes \cite{voita-titov-2020-information}, one can compare the codelength across layers of the same model or different models for a given probing task.
The edge probing method does not allow this comparison since the mixing weights do not necessarily provide an accurate estimate of the richness of linguistic knowledge within each layer.
In contrast, in MDL probing, each layer is probed separately, which gives us a direct estimate of the quality of the specific layer itself, rather than that relative to the other layers. 
MDL probing is, therefore, a better choice to have a layer-wise comparison among different models.

% ----------------------------------------------------------------
\begin{table*}[h!t!]
\begin{center}
\begin{tabular}{ l | c c | c c | c c} 
 \toprule
       & \multicolumn{2}{c|}{\textbf{BERT}} & \multicolumn{2}{c|}{\textbf{XLNet}} & \multicolumn{2}{c}{\textbf{ELECTRA}} \\
    %  \textbf{Task} & F1 & Comp. & F1 & Comp. & F1 & Comp. \\
    \textbf{Task} & F1 Score & Compression & F1 Score & Compression & F1 Score & Compression \\
     \midrule
     Deps. &    94.18 & 15.25       & 93.93 & 14.13 & \bf{94.77}     & \bf{16.15}\\
     NER &      95.61 & 16.87       & 95.51 & 15.46 & \textbf{96.07} & \textbf{16.88}\\
     SRL &      90.91 & 13.94       & 90.56 & 13.32 & \textbf{91.69} & \textbf{14.44}\\ 
     Coref. &   91.17 & \hspace{0.5em}4.58        & 91.34 & \hspace{0.5em}3.97  & \textbf{92.94} & \hspace{0.5em}\textbf{5.88}\\ 
     Rel. &     80.63 & \hspace{0.5em}3.04        & 82.07 & \hspace{0.5em}2.97  & \textbf{82.41} & \hspace{0.5em}\textbf{3.37}\\ 
\bottomrule
\end{tabular}
\end{center}
\caption{
Cross-model MDL compression and edge probing micro-averaged F1 comparison. For each task we report the highest compression achieved among a model's layers. Bold denotes the best performance on each task. In MDL probing, we employ logarithm with base 2 instead of natural logarithm in the training objective to have the codelength results in bits. Compression is the uniform codelength divided by the model's codelength. F1 scores are obtained by scalar mixing weights similar to \citet{tenney-etal-2019-bert} which are reported only for further comparison on \textbf{overall} probing performances.}

\label{fig:compression_f1}
\end{table*}
\section{Probing Pre-trained Representations}
\label{sec:pre-trained}
We begin with probing the representations of each pre-trained model on a variety of core NLP tasks, including Dependency Labeling, Named Entity Recognition, Semantic Role Labeling, Coreference Resolution, and Relation Classification (see an example for each task in Table~\ref{tab:datasets_examples}).
Following \citet{tenney2018you}, we use OntoNotes 5.0 \cite{weischedel2011ontonotes} for NER, SRL, and Coreference arc prediction, English Web Treebank portion of the Universal Dependencies \citep{silveira-etal-2014-gold} for Dependencies, and SemEval 2010 Task 8 dataset \citep{hendrickx-etal-2010-semeval} for Relation classification.
Statistics of the datasets are provided in Table~\ref{tab:dataset_stat}.

For probing evaluation, we perform MDL probing on frozen representations obtained from each model.
In order to obtain a fixed-length representation, following \citet{voita-titov-2020-information}, we project the representations to 256-dimensional vectors, and then apply a self-attention pooling on them.\footnote{For tasks with two spans, we consider separate trainable weights for projection and attention pooling.}

\subsection{Results}
\begin{figure}[h!]
\centering
    
    \begin{tikzpicture}
    \begin{groupplot}[
        group style={
            group name=my plots,
            group size=1 by 5,
            xlabels at=edge bottom,
            xticklabels at=edge bottom,
            vertical sep=0pt,
        },
        footnotesize,
        width=7.5cm,
        height=3.9cm,
        xlabel=Layer,
        xmin=-0.5, xmax=12.5,
        % ymin=1.3, % ymax=0.39,
        scaled y ticks=false,
        yticklabel style={/pgf/number format/fixed},
        legend style={font=\tiny, at={(0.55,0.03)}, anchor=south west},
        xtick={0,1,...,12},
        xticklabels={0,1,...,12},
        tickpos=left,
        ytick align=outside,
        xtick align=outside,
        xmajorgrids,
    ]
    
    % DEPS.
    \nextgroupplot[ylabel={Deps.}, ytick={6,10,14}]
    \addplot[color=blue, mark=square*, thick]
        coordinates {
        (0, 5.989924477150102)(1, 8.386370044640822)(2, 9.487929595769664)(3, 10.748528208521547)(4, 12.748485418041744)(5, 13.569377715716781)(6, 14.453910944573906)(7, 14.952917059706497)(8, 15.246287127269131)(9, 14.343860936448344)(10, 13.17113468696808)(11, 12.082044951148664)(12, 11.0602967135861)
        };
    \addplot[color=magenta, mark=*, thick]
        coordinates {
        (0, 5.012645994424057)(1, 9.390312192135971)(2, 11.370055455537084)(3, 13.18356331678955)(4, 13.984602736441632)(5, 14.001695955094082)(6, 14.13120765736814)(7, 13.568082564614256)(8, 13.298367357060982)(9, 12.717089668063217)(10, 11.627882015669158)(11, 10.422921974527624)(12, 7.2550184816664665)
        };
    \addplot[color=cyan, mark=triangle*, thick]
        coordinates {
        (0, 6.352386907961262)(1, 9.334344875153718)(2, 10.714015147176148)(3, 10.9705227005849)(4, 11.748798273472351)(5, 13.54472729805497)(6, 14.424919505386269)(7, 15.645870067283823)(8, 16.14808697928741)(9, 15.6747322065797)(10, 15.760765276751778)(11, 15.442370757205861)(12, 13.6188896527513)
        };
        
    % ENTITIES
    \nextgroupplot[ylabel={Entities}, ytick={10,12,14,16}]
    \addplot[color=blue, mark=square*, thick]
        coordinates {
        (0, 8.372565254013415)(1, 11.366185974602859)(2, 12.188023819357356)(3, 12.945903142282873)(4, 14.152524917668217)(5, 15.27891849555909)(6, 15.499068183790962)(7, 16.294200226166474)(8, 16.771681335491355)(9, 16.871782520101313)(10, 16.58757577795131)(11, 16.055141340512073)(12, 15.637886138200656)
        };
    \addplot[color=magenta, mark=*, thick]
        coordinates {
        (0, 9.328944207267993)(1, 12.143161291227347)(2, 13.562022221704957)(3, 14.972924773184698)(4, 15.46213802022247)(5, 15.304354553754353)(6, 14.850469992136262)(7, 14.294233865395412)(8, 14.203493126229828)(9, 13.72324018100344)(10, 13.357869974210578)(11, 12.774122897971667)(12, 10.841146387356961)
        };
    \addplot[color=cyan, mark=triangle*, thick]
        coordinates {
        (0, 8.722304568367164)(1, 11.742159252835869)(2, 13.549175461822353)(3, 14.792867588387368)(4, 15.252562611577305)(5, 16.770527657944793)(6, 16.883573873045034)(7, 16.602098247687856)(8, 16.762603763121326)(9, 16.456834180713518)(10, 16.287824270113134)(11, 15.466607304953136)(12, 14.099299908800212)
        };
        
    % SRL
    \nextgroupplot[ylabel={SRL}]
    \addplot[color=blue, mark=square*, thick]
        coordinates {
        (0, 7.933334369029208)(1, 9.667431760813837)(2, 10.424223486028257)(3, 11.377079768965864)(4, 12.448327851319831)(5, 13.251575623478868)(6, 13.727151197849103)(7, 13.938581899480734)(8, 13.877553809709731)(9, 13.26842139957605)(10, 12.544983514909838)(11, 12.00028501473851)(12, 11.459165017525454)
        };
    \addplot[color=magenta, mark=*, thick]
        coordinates {
        (0, 6.914117648399285)(1, 10.6968690830649)(2, 11.901093541407018)(3, 12.949318490589132)(4, 13.319018895038791)(5, 13.22715029044813)(6, 13.04913978624387)(7, 12.50248596656788)(8, 12.051979643271745)(9, 11.567376536459086)(10, 11.05699818176689)(11, 10.362882472404017)(12, 8.399671245667447)
        };
    \addplot[color=cyan, mark=triangle*, thick]
        coordinates {
        (0, 8.065308583004649)(1, 10.553684593544151)(2, 11.643494821164598)(3, 12.069152516379646)(4, 12.772982467415074)(5, 13.789968353909423)(6, 14.133070811074148)(7, 14.443409114612507)(8, 14.323144751953912)(9, 13.94136563759343)(10, 13.779231893104662)(11, 13.524284663003593)(12, 12.473229045596568)
        };
    
    % COREF.
    \nextgroupplot[ylabel={Coref.}, ytick={3,4,5,6}]
    \addplot[color=blue, mark=square*, thick]
        coordinates {
        (0, 2.735756685610581)(1, 2.964550657818625)(2, 3.110951476354563)(3, 3.2930673869860736)(4, 3.589162347374982)(5, 3.7553911857864652)(6, 3.947070982786042)(7, 4.137732917719479)(8, 4.429499067956292)(9, 4.580937916157441)(10, 4.493499174734509)(11, 4.321189798098883)(12, 4.090282004967172)
        };
    \addplot[color=magenta, mark=*, thick]
        coordinates {
        (0, 2.651067776107837)(1, 3.0287559241966284)(2, 3.3127071903851757)(3, 3.630708230725585)(4, 3.8787355390206755)(5, 3.9196874481447357)(6, 3.9514901933313813)(7, 3.9422990667696602)(8, 3.9321312509480153)(9, 3.972837770378502)(10, 3.782572335930659)(11, 3.5007213836856668)(12, 2.816302086663382)
        };
    \addplot[color=cyan, mark=triangle*, thick]
        coordinates {
        (0, 2.7263285572947558)(1, 3.057624891329783)(2, 3.2281359598592076)(3, 3.330947749614244)(4, 3.5719765941053585)(5, 4.0417022534191105)(6, 4.288777285415456)(7, 4.561138313696697)(8, 5.063234076798385)(9, 5.443943320001695)(10, 5.875174128934826)(11, 5.6840003071225444)(12, 4.757577018882047)
        };
    
    % REL.
    \nextgroupplot[ylabel={Relations}, ytick={2,3}]
    \addplot[color=blue, mark=square*, thick]
        coordinates {
        (0, 1.5065522192689418)(1, 1.796183581182723)(2, 1.9101723154950019)(3, 2.1223115070995453)(4, 2.3073083307769533)(5, 2.4816318982676737)(6, 2.5140264628885762)(7, 2.737520814809641)(8, 2.9770497178328)(9, 3.038016659981585)(10, 2.9276673763503305)(11, 2.7698616595799646)(12, 2.57035746287171)
        };
    \addplot[color=magenta, mark=*, thick]
        coordinates {
        (0, 1.6582279140252012)(1, 1.9543657104690495)(2, 2.1640558095594975)(3, 2.505549405167095)(4, 2.5906640357251596)(5, 2.6791724354890265)(6, 2.8876312907338746)(7, 2.8982049246781116)(8, 2.9513789823902443)(9, 2.9467790043807445)(10, 2.9694882985795172)(11, 2.8208127302413506)(12, 2.33472579912666)
        };
    \addplot[color=cyan, mark=triangle*, thick]
        coordinates {
        (0, 1.6295592477180592)(1, 1.8529679009683913)(2, 2.0001540212023965)(3, 2.1790051308449265)(4, 2.335425849412618)(5, 2.8014286716598993)(6, 2.8928699033517065)(7, 3.0357236163167847)(8, 3.366853658105278)(9, 3.3444956893437805)(10, 3.319735887488967)(11, 2.9787713175699526)(12, 2.5734538919442533)
        };
    \legend{BERT,XLNet,ELECTRA}
    \end{groupplot}
    \end{tikzpicture}
    \caption{MDL probing compression of BERT, XLNet, and ELECTRA across layers.}
    \label{fig:mdl_probing_layers}
\end{figure}
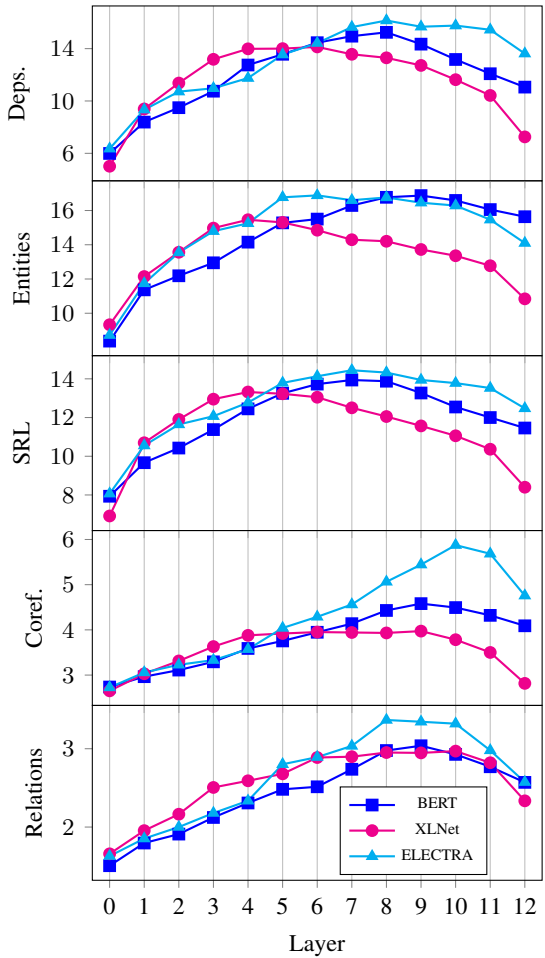
For an overall cross-model comparison, we report the MDL probe compression and edge probing F1 score results in Table \ref{fig:compression_f1}. 
For each model, the best MDL compression across different layers is reported.
ELECTRA consistently achieves the highest compression and edge probing micro-averaged F1\footnote{In relation classification, we ignore the "Other" label for calculating the F1 score.} according to both metrics. 
Both results demonstrate how well the five tasks are encoded in the models' representations during pre-training.
\nohyphens{Table~\ref{fig:compression_f1}} shows that ELECTRA can achieve the best quality in both MDL probing and edge probing compared to BERT and \nohyphens{XLNet}.
Hence, ELECTRA seems to have the best pre-training objective for incorporating linguistic knowledge among the three models. On the other hand, \nohyphens{XLNet} displays comparable results to BERT, which is interesting given the relatively better fine-tuned performance of the former in a variety of downstream tasks. 

\paragraph{Layer-wise analysis.}
Next, we use MDL probing to investigate how much linguistic knowledge is encoded in different layers in these models.
Figure \ref{fig:mdl_probing_layers} shows layer-wise MDL probing compression results of BERT, \nohyphens{XLNet}, and ELECTRA on five probing tasks. Higher compression indicates better encoding of the task. 
As can be seen, ELECTRA attains the highest compression in different layers across most tasks, especially in the deeper layers.
Notably, all models start with relatively low compressions and reach higher values in their middle layers.
An interesting behavior shared among the three models is the decrease towards the final layer, which can be attributed to their pre-training objectives. 
The main difference between the models lies in the position in which the maximum amount of linguistic knowledge is accumulated.

\begin{figure}[h!]
\centering

    \begin{tikzpicture}
      \begin{axis}[
        % title  = MDL Compression Center of Gravity,
        width=8.2cm, height=8.3cm,
        xbar,
        bar width=1.9ex,
        y axis line style = { opacity = 0 },
        % axis x line       = none,
        % tickwidth         = 1pt,
        xtick align=outside,
        yticklabel style={rotate=90,anchor=base,yshift=0.2cm},
        xmajorgrids,
        xmin=5.5, xmax=7.3,
        % enlarge y limits  = 0.9,
        % enlarge x limits  = 0.02,
        symbolic y coords = {Deps., Entities, SRL, Coref., Relations},
        nodes near coords,
        y dir=reverse,
        legend style={font=\tiny, at={(0.01,0.01)}, anchor=south west},
        legend image code/.code={%
            \draw[#1, draw=none] (0cm,-0.1cm) rectangle (0.6cm,0.1cm);
        },
        reverse legend,
      ]

    %   \addplot[color=blue] coordinates { (57727,LaTeX)(5672,Tools)(2193,Distributions)(11106,Editors)
    %   };
    % ELECTRA
      \addplot[preaction={fill, cyan}, pattern=crosshatch, pattern color=white]
      coordinates {
        (6.7045,Deps.)(6.3603,Entities)(6.3577,SRL)(6.8223,Coref.)(6.6518,Relations)
      };
    %   BERT
      \addplot[preaction={fill, blue}, pattern=dots, pattern color=white] coordinates {
        (6.5171,Deps.)(6.5460,Entities)(6.3240,SRL)(6.5332,Coref.)(6.6211,Relations)
      };
   %   XLNET
      \addplot[preaction={fill, magenta}, pattern=grid, pattern color=white] coordinates {
        (6.1096,Deps.)(6.0236,Entities)(5.9760,SRL)(6.1378,Coref.)(6.4159,Relations)
      };
      
      \legend{ELECTRA, BERT, XLNet}
      \end{axis}
    \end{tikzpicture}

    \caption{Comparison of the ~MDL probing compression center of gravity in BERT, XLNet, and \nohyphens{ELECTRA}.}
    \label{fig:mdl_cg}
\end{figure}
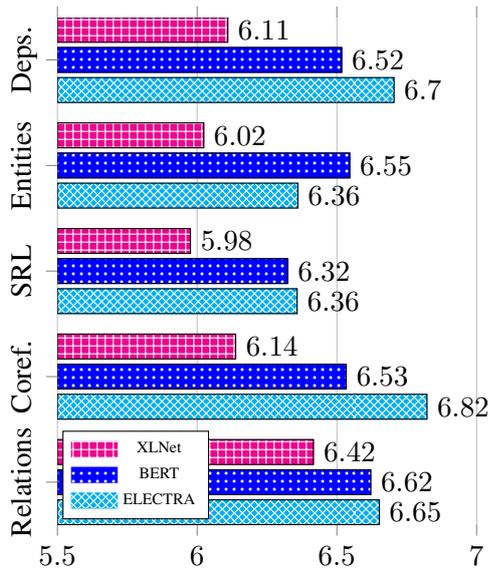

To better demonstrate the layer that most captures each task, we compute the center of gravity following \citet{tenney-etal-2019-bert}. The only difference is that we apply it on MDL probing compression, instead of scalar mixing weights, defined as:
\begin{equation}
\bar{E}_{\mathbf{c}}[\ell] = \frac{\sum_{\ell=0}^L \ell \cdot \mathbf{c}^{(\ell)}}{\sum_{\ell=0}^L \mathbf{c}^{(\ell)}}
\label{eq:cg}
\vspace{+0.8ex}
\end{equation}
where $\mathbf{c}^{(\ell)}$ is the compression score of layer $\ell$.
Figure \ref{fig:mdl_cg} shows the center of gravity of compression. The most noticeable distinction among models is that \nohyphens{XLNet's} linguistic knowledge is concentrated in earlier layers than BERT, while ELECTRA's knowledge is mostly accumulated in deeper layers. 

We hypothesize that the difficulty of the objectives has a direct effect on $\bar{E}_{\mathbf{c}}$, which indicates the expected position with the most encoded linguistic knowledge. 
In particular, recovering input tokens in the final layers of the model in the pre-training objective of BERT and \nohyphens{XLNet} is a surface task. 
Some of the linguistic knowledge might diminish in the final layers since highly contextualized representations have to be transformed into a less contextualized level to predict the original inputs \cite{voita2019bottom}.
Whereas the pre-training objective in \nohyphens{ELECTRA} might be considered as a more semantic task, in which detecting replaced tokens requires more context-aware representations.

% ------------------------------------------------------------

\section{Probing Fine-tuned Representations}
After observing that the amount and distribution of encoded linguistic knowledge can differ in these models, we aim to investigate how this information might be affected after the fine-tuning process. 
To this end, we repeated MDL probing as described in Section \ref{sec:pre-trained} on the fine-tuned representations. 
Next, to validate our results, we will expand our experiments to complementary analyses and measure the final quality of these representations on downstream tasks at different layers.
We opted for the MNLI \cite{williams-etal-2018-broad} dataset for fine-tuning all models. 
We used the same hyper-parameters for fine-tuning all three models: 32 as the batch size, max length of 128, the learning rate of 2e-5, and five epochs of training.

\begin{figure}[t]
\centering
\includegraphics[width=0.44\textwidth]{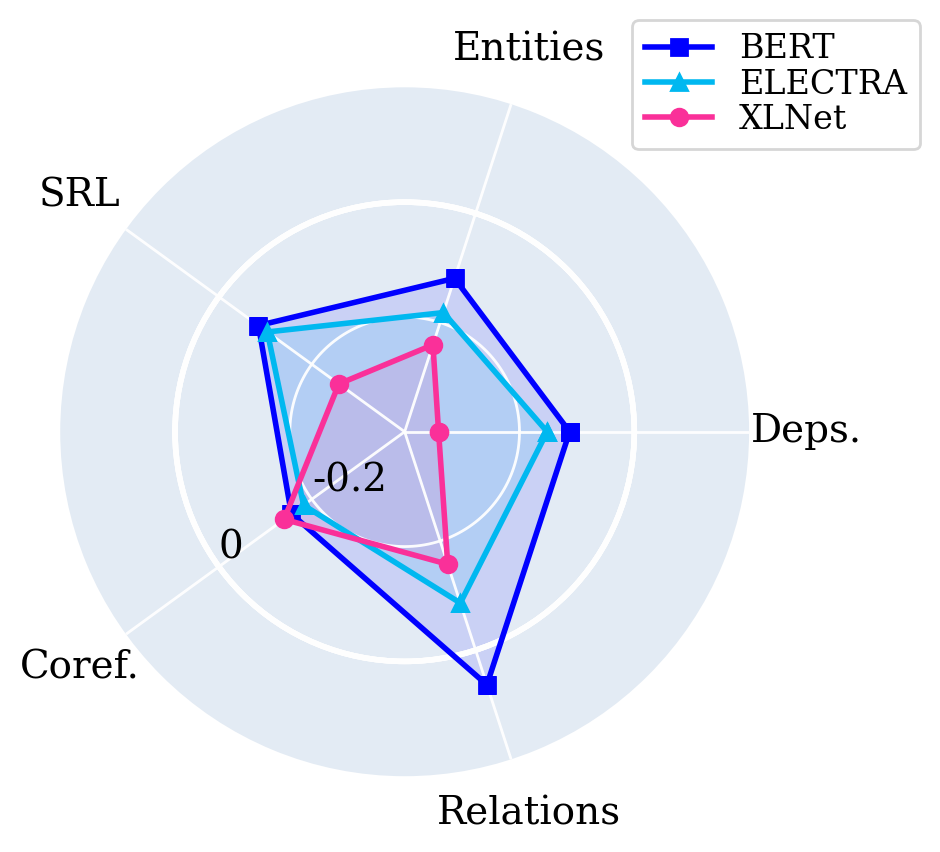}
\caption{The change in centers of gravity after fine-tuning on MNLI dataset in five linguistic tasks.}

\label{fig:dif_cg_radar}
\end{figure}

\paragraph{Center of Gravity in fine-tuned models.}
Figure \ref{fig:dif_cg_radar} demonstrates the difference in the average layer that most captures the information related to a particular task between the pre-trained and the fine-tuned models. 
We measure the difference of the two centers of gravity to evaluate the extent to which the concentration of knowledge ($\bar{E}_{\mathbf{c}}$) shifts for each model on a specific probing task after fine-tuning:
\begin{equation}
    \Delta \bar{E}_{\mathbf{c}} = \bar{E}_{\mathbf{c}}^{finetuned} - \bar{E}_{\mathbf{c}}^{pretrained}
\end{equation}
First, we show that the concentration of information in fine-tuned models is usually in earlier layers compared to the pre-trained models. This can be attributed to the significant loss of linguistic knowledge in the final layers of fine-tuned models in favor of the specific information of the fine-tuning task.
We show that \nohyphens{XLNet} in most tasks falls back to earlier layers than the two other models because it forgets the most linguistic knowledge in the final layers. This suggests that \nohyphens{XLNet} is going through a more extensive change in its representations which we investigate in the following sections. We also indicate that \nohyphens{ELECTRA} is falling back more than BERT. We hypothesize that since \nohyphens{ELECTRA} focuses its most capable representations in the final layers, and those are the layers that change the most in fine-tuning, its center of gravity changes more than BERT.
Full MDL compression and codelength results are reported in the appendix (Table \ref{tab:compressions_appendix}).

\paragraph{Global RSA.}
After fine-tuning each model, we leverage Representational Similarity Analysis (RSA) to investigate the overall amount of changes in the representations of each layer.
RSA is a technique borrowed from neuroscience \cite{neuroscience-rsa} which is used for comparing two different representation spaces. 
To be specific, we sampled 5000 English sentences from OPUS dataset \citep{tiedemann-2012-parallel} and used Global RSA\footnote{Using the average pooled representations as stimuli instead of individual tokens.}, introduced by \citet{chrupala-etal-2020-analyzing}. This is a better choice since we used average pooling in our fine-tuning, discussed next. As a measure of intra- and inter-space similarities, we used cosine similarity and Pearson correlation, respectively.

\begin{figure}[t]
\centering
    \begin{tikzpicture}
    \begin{axis}[
        width=0.48\textwidth,
        height=7.5cm,
        title style={yshift=-1.4ex},
        xlabel={Layer},
        ylabel={RSA Similarity},
        title style={font=\small},
        label style={font=\small},
        legend style={font=\small},
        % xmin=0, %xmax=100,
        ymin=-0.07, ymax=1.03,
        xtick={0,2,4,6,8,10,12},
        legend pos=south west,
        grid=both,
        % ymajorgrids=true,
        % grid style=dashed,
        % line width=2pt,
    ]
    
    % BERT
    \addplot[
        color=blue,
        mark=square*,
        thick,
        error bars/.cd, y dir=both, y explicit,
        ]
        coordinates {
        (0, 0.9985372357898288) +=(0, 0.0005405611462063753) -=(0, 0.0005345874362521208)
        (1, 0.9813912643326653) +=(1, 0.0013218985663520089) -=(1, 0.0017472041977776298)
        (2, 0.9600447085168626) +=(2, 0.003155741426679848) -=(2, 0.005348708894517662)
        (3, 0.9598525232738919) +=(3, 0.004454606109195236) -=(3, 0.004855281776852127)
        (4, 0.9573240743743049) +=(4, 0.0053334368599785575) -=(4, 0.004841374026404499)
        (5, 0.9586590263578627) +=(5, 0.0035865571763780135) -=(5, 0.0031918618414137345)
        (6, 0.9505831797917684) +=(6, 0.0012815395991007117) -=(6, 0.001575013001759884)
        (7, 0.9301924440595839) +=(7, 0.004916992452409508) -=(7, 0.009052786562177895)
        (8, 0.9061446322335137) +=(8, 0.008512722121344685) -=(8, 0.015669478310479046)
        (9, 0.8638241092363993) +=(9, 0.016289790471394894) -=(9, 0.009333531061808231)
        (10, 0.8501187033123441) +=(10, 0.012582414680057052) -=(10, 0.016041464275784012)
        (11, 0.8651748365826077) +=(11, 0.013978295856051925) -=(11, 0.008465773529476639)
        (12, 0.634150332874722) +=(12, 0.034580104880862716) -=(12, 0.05510760015911531)
        };

    % XLNet
    \addplot[
        color=magenta,
        mark=*,
        thick,
        error bars/.cd, y dir=both, y explicit,
        ]
        coordinates {
        (0, 0.9988192319869995) +=(0, 5.644559860229492e-05) -=(0, 5.6684017181396484e-05)
        (1, 0.6579458978441026) +=(1, 0.028779003355238197) -=(1, 0.02620169189241195)
        (2, 0.7469193802939521) +=(2, 0.007916523350609661) -=(2, 0.009457753764258503)
        (3, 0.6919508510165744) +=(3, 0.01386440462536287) -=(3, 0.013748877578311447)
        (4, 0.842812180519104) +=(4, 0.005322515964508057) -=(4, 0.010984420776367188)
        (5, 0.7992253767119514) +=(5, 0.00443817509545219) -=(5, 0.00482325421439278)
        (6, 0.6373478306664361) +=(6, 0.007544921504126667) -=(6, 0.007883383168114544)
        (7, 0.6236078010665046) +=(7, 0.006202532185448528) -=(7, 0.00879816876517403)
        (8, 0.49740247593985665) +=(8, 0.018338574303521038) -=(8, 0.017501549588309406)
        (9, 0.5730700294176737) +=(9, 0.022423168023427364) -=(9, 0.02933601538340247)
        (10, 0.45843274063534206) +=(10, 0.029918723636203348) -=(10, 0.049098468489117086)
        (11, 0.03868900795431626) +=(11, 0.08080107907800185) -=(11, 0.05019729209096921)
        (12, -0.022069533277923863) +=(12, 0.01984699039409558) -=(12, 0.019045400510852534)
        };
    
    % ELECTRA
    \addplot[color=cyan, mark=triangle*, thick, error bars/.cd, y dir=both, y explicit,]
        coordinates {
        (0, 0.9984803199768066) +=(0, 0.00029224157333374023) -=(0, 0.00025260448455810547)
        (1, 0.9856415523423089) +=(1, 0.0012720690833197823) -=(1, 0.0012938512696160087)
        (2, 0.971453845500946) +=(2, 0.004319190979003906) -=(2, 0.005438089370727539)
        (3, 0.9685534238815308) +=(3, 0.0032451748847961426) -=(3, 0.004460334777832031)
        (4, 0.965825531217787) +=(4, 0.002281513479020836) -=(4, 0.0032797786924574357)
        (5, 0.9513216416041056) +=(5, 0.006187756856282589) -=(5, 0.005044261614481571)
        (6, 0.9656561877992418) +=(6, 0.003808233473036049) -=(6, 0.0038473274972703475)
        (7, 0.9604023959901598) +=(7, 0.0020574066374037248) -=(7, 0.0026774075296189803)
        (8, 0.9451854361428155) +=(8, 0.002845035658942341) -=(8, 0.004086150063408733)
        (9, 0.9168940782546997) +=(9, 0.009619176387786865) -=(9, 0.005760610103607178)
        (10, 0.913231286737654) +=(10, 0.01011386182573104) -=(10, 0.009676012727949379)
        (11, 0.872679180569119) +=(11, 0.020121090941958908) -=(11, 0.0156389541096158)
        (12, 0.4872358938058217) +=(12, 0.04745269815127057) -=(12, 0.04959203799565631)
        };
    
    % ERROR Regions
    % BERT
    \addplot [name path=upper_bert, fill=none, draw=none]  
    coordinates {
        (0, 0.9990777969360352)(1, 0.9827131628990173)(2, 0.9632004499435425)(3, 0.9643071293830872)(4, 0.9626575112342834)(5, 0.9622455835342407)(6, 0.9518647193908691)(7, 0.9351094365119934)(8, 0.9146573543548584)(9, 0.8801138997077942)(10, 0.8627011179924011)(11, 0.8791531324386597)(12, 0.6687304377555847)
    };
    \addplot [name path=lower_bert, fill=none, draw=none]  
    coordinates {
        (0, 0.9980026483535767)(1, 0.9796440601348877)(2, 0.954695999622345)(3, 0.9549972414970398)(4, 0.9524827003479004)(5, 0.955467164516449)(6, 0.9490081667900085)(7, 0.921139657497406)(8, 0.8904751539230347)(9, 0.8544905781745911)(10, 0.8340772390365601)(11, 0.8567090630531311)(12, 0.5790427327156067)
    };
    \addplot[blue!40] fill between[of=lower_bert and upper_bert];
    
    % XLNet
    \addplot [name path=upper_xlnet, fill=none, draw=none]  
    coordinates {
        (0, 0.9988756775856018)(1, 0.6867249011993408)(2, 0.7548359036445618)(3, 0.7058152556419373)(4, 0.8481346964836121)(5, 0.8036635518074036)(6, 0.6448927521705627)(7, 0.6298103332519531)(8, 0.5157410502433777)(9, 0.5954931974411011)(10, 0.4883514642715454)(11, 0.11949008703231812)(12, -0.0022225428838282824)
    };
    \addplot [name path=lower_xlnet, fill=none, draw=none]  
    coordinates {
        (0, 0.9987625479698181)(1, 0.6317442059516907)(2, 0.7374616265296936)(3, 0.6782019734382629)(4, 0.8318277597427368)(5, 0.7944021224975586)(6, 0.6294644474983215)(7, 0.6148096323013306)(8, 0.47990092635154724)(9, 0.5437340140342712)(10, 0.409334272146225)(11, -0.011508284136652946)(12, -0.0411149337887764)
    };
    \addplot[magenta!40] fill between[of=lower_xlnet and upper_xlnet];
    
    % ELECTRA
    \addplot [name path=upper_electra, fill=none, draw=none]  
    coordinates {
        (0, 0.9987725615501404)(1, 0.9869136214256287)(2, 0.97577303647995)(3, 0.9717985987663269)(4, 0.9681070446968079)(5, 0.9575093984603882)(6, 0.9694644212722778)(7, 0.9624598026275635)(8, 0.9480304718017578)(9, 0.9265132546424866)(10, 0.923345148563385)(11, 0.8928002715110779)(12, 0.5346885919570923)
    };
    \addplot [name path=lower_electra, fill=none, draw=none]  
    coordinates {
        (0, 0.9982277154922485)(1, 0.9843477010726929)(2, 0.9660157561302185)(3, 0.9640930891036987)(4, 0.9625457525253296)(5, 0.946277379989624)(6, 0.9618088603019714)(7, 0.9577249884605408)(8, 0.9410992860794067)(9, 0.9111334681510925)(10, 0.9035552740097046)(11, 0.8570402264595032)(12, 0.4376438558101654)
    };
    \addplot[cyan!40] fill between[of=lower_electra and upper_electra];
    
    \legend{BERT, XLNET, ELECTRA}
        
    \end{axis}
    \end{tikzpicture}
    
    \caption{Similarity of the representations of \nohyphens{BERT}, \nohyphens{XLNet}, and \nohyphens{ELECTRA}, before and after fine-tuning on MNLI dataset. Additional plots for CoLA and SST-2 are provided in the appendix (Figures ~\ref{fig:rsa_cola} and ~\ref{fig:rsa_sst2}).}
    \label{fig:rsa_mnli}
\end{figure}
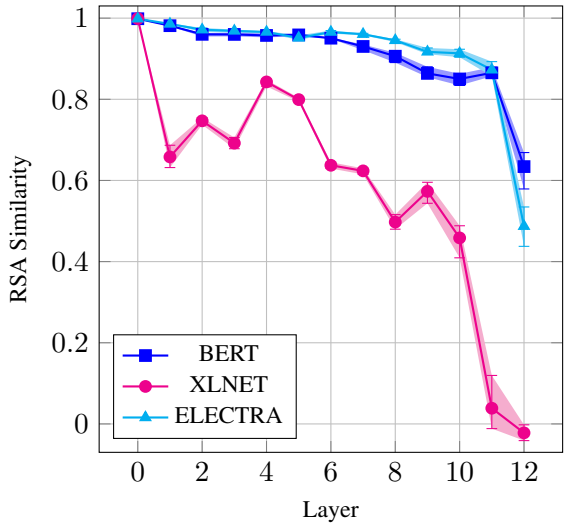
Figure \ref{fig:rsa_mnli} shows the results of the RSA measure applied to the representations of our models. We observe that \nohyphens{XLNet} has changed drastically during fine-tuning, while in BERT and \nohyphens{ELECTRA}, only the top layers are primarily affected. 
We also see that BERT shows a conservative pattern in the fine-tuning process which is consistent with findings of \citet{merchant-etal-2020-happens}.
As seen in Figure \ref{fig:mdl_probing_layers}, we hypothesize that higher layers in \nohyphens{XLNet} are more open to change since they have relatively less specific knowledge. On the contrary, \nohyphens{ELECTRA}, which enjoys more linguistic information in its representations, especially in the higher layers, does not need to change very much.

\paragraph{Quality of the representations for downstream tasks.}
With the observations on RSA curves, we were interested in knowing the impact of the extent of these changes on downstream performance.
To this end, we evaluated the quality of the representations for downstream tasks in both pre-trained and fine-tuned models.
We trained separate classifiers on the unweighted average of representations for each layer.\footnote{Unlike the other two models, BERT involves [CLS] token for NSP objective during pre-training step.
Hence, to have a fair comparison between our fine-tuning and feature extraction experiments, we opted for the mean pooling strategy consistently throughout the paper.} We used  Adam optimizer with a learning rate of 5e-4, and binary cross-entropy loss function for optimization.
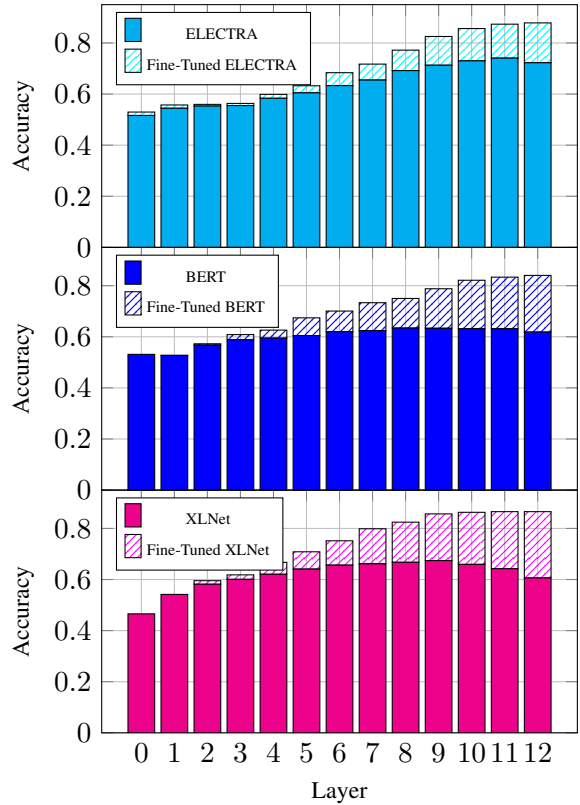
\begin{figure}[t]
\centering
    \begin{tikzpicture}
    \begin{groupplot}[
        group style={
            group name=my plots,
            group size=1 by 3,
            xlabels at=edge bottom,
            xticklabels at=edge bottom,
            vertical sep=0pt,
        },
        ybar stacked, 
        width=0.49\textwidth,
        height=4.8cm,
        xlabel={Layer},
        ylabel={Accuracy},
        title style={font=\small},
        label style={font=\small},
        legend style={font=\tiny},
        % xmin=0, %xmax=100,
        ymin=0, ymax=0.95,
        xtick={0,...,12},
        % xticklabel interval boundaries,
        % ytick={0,1,2,3,4,5,6,7,8,9,10,11,12},
        legend pos=north west,
        grid=both,
    ]
    
    % ELECTRA
    \nextgroupplot[]
    \addplot[
        ybar, fill=cyan,
        ]
        coordinates {
            (0, 0.5162506367804381)(1, 0.5444727457972491)(2, 0.5530310748853795)(3, 0.5550687722873153)(4, 0.58349465104432)(5, 0.6047885888945491)(6, 0.633112582781457)(7, 0.6548140601120733)(8, 0.6917982679572083)(9, 0.7128884360672441)(10, 0.7300050942435048)(11, 0.7407030056036679)(12, 0.7225674987264391)
        };
    \addplot[
        ybar, pattern=north east lines, pattern color=cyan
        ]
        coordinates {
            (0, 0.013346917982679507)(1, 0.012837493632195551)(2, 0.0066225165562914245)(3, 0.00835455934793683)(4, 0.01538461538461533)(5, 0.027712684666327148)(6, 0.050534895568008165)(7, 0.062455425369332684)(8, 0.08008150789607749)(9, 0.11248089658685678)(10, 0.12633723892002047)(11, 0.132857870606215)(12, 0.15598573611818645)
        };
    \legend{ELECTRA, Fine-Tuned ELECTRA}
    
    % BERT
    \nextgroupplot[]
    \addplot[
        ybar, fill=blue,
        ]
        coordinates {
            (0, 0.5313295975547632)(1, 0.5269485481406011)(2, 0.5674987264391238)(3, 0.5884870096790626)(4, 0.5949057564951604)(5, 0.604890473764646)(6, 0.6197656647987774)(7, 0.624350483953133)(8, 0.6342333163525217)(9, 0.6334182373917473)(10, 0.6316861946001019)(11, 0.6314824248599084)(12, 0.6185430463576159)
        };
    \addplot[
        ybar, pattern=north east lines, pattern color=blue
        ]
        coordinates {
            (0, -0.007539480387162567)(1, 0.0004075394803871868)(2, 0.005196128374936326)(3, 0.02007131940906781)(4, 0.03127865511971473)(5, 0.06928171166581754)(6, 0.08089658685685175)(7, 0.10901681100356597)(8, 0.11584309730005093)(9, 0.15455934793683146)(10, 0.1896077432501273)(11, 0.201935812531839)(12, 0.2217014773306164)
        };
    \legend{BERT, Fine-Tuned BERT}
    
    % XLNet
    \nextgroupplot[]
    \addplot[
        ybar, fill=magenta,
        ]
        coordinates {
            (0, 0.46561385634233315)(1, 0.5417218543046357)(2, 0.5815588385124809)(3, 0.6007131940906776)(4, 0.6205807437595517)(5, 0.6408558329088131)(6, 0.6563423331635252)(7, 0.6615384615384615)(8, 0.6674477840040753)(9, 0.67396841569027)(10, 0.6592969943963322)(11, 0.6422822210901681)(12, 0.6061130922058074)
        };
    \addplot[
        ybar, pattern=north east lines, pattern color=magenta
        ]
        coordinates {
            (0, -0.0016301579215486361)(1, -0.012633723892002013)(2, 0.013856342333163463)(3, 0.017524197656647922)(4, 0.04666327050433006)(5, 0.0676515537442689)(6, 0.09526235354049928)(7, 0.1371370351502802)(8, 0.15700458481915436)(9, 0.18278145695364234)(10, 0.20376974019358118)(11, 0.22333163525216504)(12, 0.2595007641365257)
        };
    \legend{XLNet, Fine-Tuned XLNet}

    \end{groupplot}
    \end{tikzpicture}
    
    \caption{Layer-wise comparison of the performance scores on MNLI dataset across the representations of the pre-trained and fine-tuned models.}
    \label{fig:feature_mnli}
\end{figure}

Results are shown in Figure \ref{fig:feature_mnli}.
Based on the performance scores for pre-trained representations, we observe that \nohyphens{XLNet} encodes most essential information for the downstream task in the shallower layers, BERT in the middle ones, and \nohyphens{ELECTRA} in the deeper layers. 
Interestingly, these patterns are well aligned with the MDL probing curves in Figure \ref{fig:mdl_probing_layers} which indicates that pre-trained representations with more linguistic knowledge are more suitable for downstream tasks.
We also report our results on CoLA \cite{warstadt-etal-2019-neural} and SST-2 \cite{socher-etal-2013-recursive-sst2} datasets in Figures \ref{fig:feature_cola} and \ref{fig:feature_sst2}.
Our results show that the observations are consistent across these downstream tasks.
In addition, we show that \nohyphens{XLNet} significantly improves performance in its second half of layers, while \nohyphens{ELECTRA} undergoes smaller adjustments. 
We observe that, before fine-tuning, the last layers of \nohyphens{XLNet} have fairly similar or even lower performance than BERT. 
However, when fine-tuned, \nohyphens{XLNet} compensates for the performance deficit by injecting more task-specific information in those layers, helping the model to outperform BERT.
Finally, we demonstrate that the changes in layers and their extent are similar to what we saw in the RSA results in Figure \ref{fig:rsa_mnli}, which indicates that the changes in RSA were actually made to achieve higher quality in the fine-tuning task.

% RELATED WORK -----------------------------------------
\section{Related Work}

While extensive research has been devoted to probing BERT \cite{hewitt-manning-2019-structural, tenney-etal-2019-bert, merchant-etal-2020-happens}, other popular models, such as \nohyphens{XLNet} and \nohyphens{ELECTRA} which have significant differences in their pre-training objectives, are less thoroughly investigated.
Only a few probing studies similar to our work exist which cover other models within the family.
\citet{mosbach-etal-2020-interplay} mostly focused on the impact of pooling strategy in probing sentence-level tasks. Based on the accuracy, they found that fine-tuning can affect the linguistic knowledge encoded in the representations.
Moreover, \citet{durrani-etal-2020-analyzing} investigated the distribution of linguistic knowledge across individual neurons. In particular, they found that neurons in XLNet are more localized in encoding individual linguistic information compared to BERT, where neurons are shared across multiple properties.
By adopting the method of \citet{hewitt-manning-2019-structural}, \citet{aspillaga-etal-2021-inspecting} investigated whether pre-trained language models encode semantic information, for instance by checking their representations against the lexico-semantic structure of WordNet \cite{miller-1994-wordnet}.

The above studies mainly rely on the accuracy metric for their probing evaluation, which is recently shown to fail in adequately reflecting the differences among representations  \cite{voita-titov-2020-information}. To our knowledge, this is the first time that an information-theoretic probing method is employed for conducting a cross-model and layer-wise analytical study.

\section{Conclusions}
In this paper, we aimed to extend probing studies on BERT to the other models in the family to investigate how training objectives and architectural choices would affect the resulting representations and the linguistic knowledge encoded in them.
To this end, we leveraged MDL probing method, which has recently proven to provide more reliable and informative results when compared with conventional probes. To the best of our knowledge, this is the first time MDL probing has been employed to analyze such state-of-the-art pre-trained language models as BERT.
By probing three state-of-the-art language models, i.e., BERT, \nohyphens{XLNet}, and \nohyphens{ELECTRA}, we found considerable differences in the extent and distribution of the core linguistic knowledge in their representations.
Specifically, we demonstrate that \nohyphens{XLNet} accumulates linguistic knowledge in the earlier layers than BERT, whereas that of \nohyphens{ELECTRA} is mainly in the final layers.

Moreover, from probing and employing RSA similarity measure on fine-tuned models, we illustrate that \nohyphens{XLNet} is more susceptible to forgetting linguistic knowledge in final layers and undergoes substantial adjustments to its representations when compared to the other models. Based on differential downstream performance observations for before and after fine-tuning, we confirm that the changes in representations are proportional to the provided gain in the downstream task, which consequently indicates that \nohyphens{XLNet} injects more information during fine-tuning into its representations than the two other models.

In summary, through probing and measurement tools, we demonstrate that BERT's derivative models, especially those with different objectives and structural choices, express different behaviors in their representations. We hope our analysis helps make more informed choices in the selection and fine-tuning of these state-of-the-art models.
%  ------------------------------------------------------------------
\newpage
\section*{Acknowledgments}
Our work is in part supported by Tehran Institute for Advanced Studies (TeIAS), Khatam University.
% The acknowledgments should go immediately before the references. Do not number the acknowledgments section.
% Do not include this section when submitting your paper for review.
\bibliographystyle{acl_natbib}
% \bibliography{anthology,emnlp2020}
\bibliography{emnlp2020}

\clearpage
\appendix
\counterwithin{figure}{section}
\counterwithin{table}{section}

\section{Appendices}
\label{sec:appendix}

\subsection{Edge Probing Normalized Mixing Weights Results}
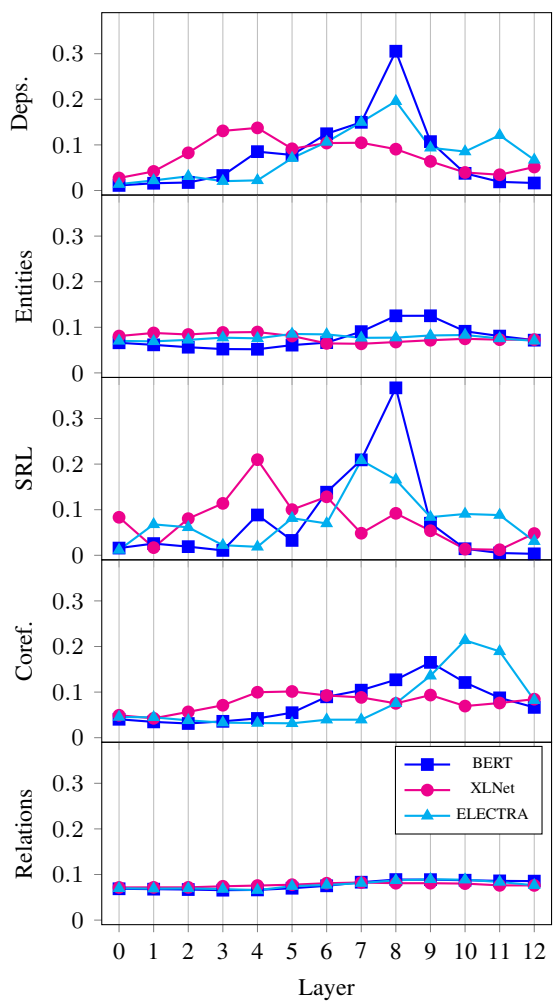
\begin{figure}[h!]
\centering
    
    \begin{tikzpicture}
    \begin{groupplot}[
        group style={
            group name=my plots,
            group size=1 by 5,
            xlabels at=edge bottom,
            xticklabels at=edge bottom,
            vertical sep=0pt,
        },
        footnotesize,
        width=7.5cm,
        height=4cm,
        xlabel=Layer,
        xmin=-0.5, xmax=12.5,
        ymin=-0.01, ymax=0.39,
        scaled y ticks=false,
        yticklabel style={/pgf/number format/fixed},
        legend style={font=\tiny},
        xtick={0,1,...,12},
        xticklabels={0,1,...,12},
        tickpos=left,
        ytick align=outside,
        xtick align=outside,
        xmajorgrids,
    ]

    % DEPS.
    \nextgroupplot[ylabel={Deps.}]
    \addplot[color=blue, mark=square*, thick]
        coordinates {
        (0, 0.01110097300261259)(1, 0.015897108241915703)(2, 0.017511488869786263)(3, 0.03283638507127762)(4, 0.08519946783781052)(5, 0.0779494121670723)(6, 0.12441803514957428)(7, 0.14947983622550964)(8, 0.30542123317718506)(9, 0.10717909038066864)(10, 0.03761071339249611)(11, 0.018940705806016922)(12, 0.01645551435649395)
        };
    \addplot[color=magenta, mark=*, thick]
        coordinates {
        (0, 0.027422355487942696)(1, 0.04182126000523567)(2, 0.08259018510580063)(3, 0.13070809841156006)(4, 0.13727715611457825)(5, 0.09146849066019058)(6, 0.10401353985071182)(7, 0.10468738526105881)(8, 0.0906473845243454)(9, 0.06382875144481659)(10, 0.03952091187238693)(11, 0.03451017662882805)(12, 0.05150427296757698)
        };
    \addplot[color=cyan, mark=triangle*, thick]
        coordinates {
        (0, 0.013958338648080826)(1, 0.02209421806037426)(2, 0.03115418180823326)(3, 0.020382113754749298)(4, 0.022251149639487267)(5, 0.07088447362184525)(6, 0.10635042190551758)(7, 0.14987851679325104)(8, 0.19569894671440125)(9, 0.09368050843477249)(10, 0.08539170771837234)(11, 0.12111695855855942)(12, 0.06715849041938782)
        };
        
    % ENTITIES
    \nextgroupplot[ylabel={Entities}]
    \addplot[color=blue, mark=square*, thick]
        coordinates {
        (0, 0.06605124473571777)(1, 0.061489567160606384)(2, 0.05632639676332474)(3, 0.05225670337677002)(4, 0.05186547338962555)(5, 0.06089966744184494)(6, 0.0664183646440506)(7, 0.09048637747764587)(8, 0.12518957257270813)(9, 0.12524844706058502)(10, 0.09126192331314087)(11, 0.08073943853378296)(12, 0.07176687568426132)
        };
    \addplot[color=magenta, mark=*, thick]
        coordinates {
        (0, 0.08080773055553436)(1, 0.08758408576250076)(2, 0.08419059962034225)(3, 0.0886889174580574)(4, 0.08959078788757324)(5, 0.08091078698635101)(6, 0.06470795720815659)(7, 0.06376826018095016)(8, 0.06768801063299179)(9, 0.0716153085231781)(10, 0.07474119961261749)(11, 0.07284514605998993)(12, 0.0728612095117569)
        };
    \addplot[color=cyan, mark=triangle*, thick]
        coordinates {
        (0, 0.07018368691205978)(1, 0.06913457065820694)(2, 0.0724857747554779)(3, 0.0772533044219017)(4, 0.07578060775995255)(5, 0.08538519591093063)(6, 0.0844438225030899)(7, 0.07688000798225403)(8, 0.07733183354139328)(9, 0.08210945129394531)(10, 0.08307788521051407)(11, 0.07552371919155121)(12, 0.07041020691394806)
        };
        
    % SRL
    \nextgroupplot[ylabel={SRL}]
    \addplot[color=blue, mark=square*, thick]
        coordinates {
        (0, 0.015841782093048096)(1, 0.02569570019841194)(2, 0.018965426832437515)(3, 0.010712417773902416)(4, 0.08840827643871307)(5, 0.032600462436676025)(6, 0.13837993144989014)(7, 0.2092597633600235)(8, 0.367209792137146)(9, 0.07035167515277863)(10, 0.01435608696192503)(11, 0.0050307405181229115)(12, 0.0031880075111985207)
        };
    \addplot[color=magenta, mark=*, thick]
        coordinates {
        (0, 0.08334410935640335)(1, 0.016788357868790627)(2, 0.08032466471195221)(3, 0.11382485181093216)(4, 0.20967906713485718)(5, 0.10013879090547562)(6, 0.1282282918691635)(7, 0.04837750643491745)(8, 0.09201786667108536)(9, 0.053885601460933685)(10, 0.013507617637515068)(11, 0.012170340865850449)(12, 0.04771291837096214)
        };
    \addplot[color=cyan, mark=triangle*, thick]
        coordinates {
        (0, 0.01159778330475092)(1, 0.06783073395490646)(2, 0.06133569777011871)(3, 0.022152289748191833)(4, 0.018549304455518723)(5, 0.08131511509418488)(6, 0.06981422752141953)(7, 0.2082746922969818)(8, 0.16574017703533173)(9, 0.08371414989233017)(10, 0.09063400328159332)(11, 0.08818710595369339)(12, 0.030854705721139908)
        };
    
    % COREF.
    \nextgroupplot[ylabel={Coref.}]
    \addplot[color=blue, mark=square*, thick]
        coordinates {
        (0, 0.04039392247796059)(1, 0.034652337431907654)(2, 0.03113841637969017)(3, 0.03590519726276398)(4, 0.0422041118144989)(5, 0.05478684604167938)(6, 0.08963853865861893)(7, 0.10425002872943878)(8, 0.12700827419757843)(9, 0.16516438126564026)(10, 0.12103620171546936)(11, 0.0873662456870079)(12, 0.06645549833774567)
        };
    \addplot[color=magenta, mark=*, thick]
        coordinates {
        (0, 0.04918617382645607)(1, 0.04248877242207527)(2, 0.056563254445791245)(3, 0.07115577906370163)(4, 0.09967996180057526)(5, 0.10138405859470367)(6, 0.09260156750679016)(7, 0.0883386954665184)(8, 0.07526843249797821)(9, 0.09346847236156464)(10, 0.06932198256254196)(11, 0.07608797401189804)(12, 0.08445482701063156)
        };
    \addplot[color=cyan, mark=triangle*, thick]
        coordinates {
        (0, 0.045400023460388184)(1, 0.04466709867119789)(2, 0.03805654123425484)(3, 0.03303801268339157)(4, 0.03214704245328903)(5, 0.03152944892644882)(6, 0.03947361186146736)(7, 0.03950994461774826)(8, 0.07534262537956238)(9, 0.13579480350017548)(10, 0.21350930631160736)(11, 0.1892990916967392)(12, 0.08223243057727814)
        };
    
    % REL.
    \nextgroupplot[ylabel={Relations}]
    \addplot[color=blue, mark=square*, thick]
        coordinates {
        (0, 0.06868410855531693)(1, 0.06783915311098099)(2, 0.06701570749282837)(3, 0.06560598313808441)(4, 0.06617707759141922)(5, 0.06986843049526215)(6, 0.0752013623714447)(7, 0.08309426158666611)(8, 0.08888376504182816)(9, 0.08871982991695404)(10, 0.08746067434549332)(11, 0.0860004797577858)(12, 0.08544918894767761)
        };
    \addplot[color=magenta, mark=*, thick]
        coordinates {
        (0, 0.07160694152116776)(1, 0.07176563888788223)(2, 0.07179750502109528)(3, 0.07414790987968445)(4, 0.07573571056127548)(5, 0.07762087136507034)(6, 0.0808178260922432)(7, 0.08247406035661697)(8, 0.08084630966186523)(9, 0.08091741800308228)(10, 0.0800359919667244)(11, 0.07618522644042969)(12, 0.07604863494634628)
        };
    \addplot[color=cyan, mark=triangle*, thick]
        coordinates {
        (0, 0.0697953924536705)(1, 0.06901872903108597)(2, 0.06953330338001251)(3, 0.06804271787405014)(4, 0.06568960100412369)(5, 0.07444634288549423)(6, 0.07700683921575546)(7, 0.08111453056335449)(8, 0.08732471615076065)(9, 0.08941858261823654)(10, 0.08857836574316025)(11, 0.08359372615814209)(12, 0.07643712311983109)
        };
    \legend{BERT,XLNet,ELECTRA}
    
    \end{groupplot}
    \end{tikzpicture}

    \caption{Edge probing mixing weights results for BERT, XLNet, and ELECTRA. This is a modified version of mixing weights where we normalize the representations before applying mixing weights to eliminate the norms disparity effect.}
    \label{fig:edge_probing_layers}
\end{figure}

\subsection{Results for CoLA Dataset}
Figures \ref{fig:feature_cola} and \ref{fig:rsa_cola} report the performance and RSA results before and after fine-tuning on CoLA dataset.
The results are consistent with figure \ref{fig:feature_mnli}.
\begin{figure}[h]
\centering
    \begin{tikzpicture}
    \begin{groupplot}[
        group style={
            group name=my plots,
            group size=1 by 3,
            xlabels at=edge bottom,
            xticklabels at=edge bottom,
            vertical sep=0pt,
        },
        ybar stacked, 
        width=0.49\textwidth,
        height=5cm,
        xlabel={Layer},
        ylabel={Matthews Correlation},
        title style={font=\small},
        label style={font=\small},
        legend style={font=\tiny},
        % xmin=0, %xmax=100,
        ymin=0, ymax=0.75,
        xtick={0,...,12},
        % xticklabel interval boundaries,
        % ytick={0,1,2,3,4,5,6,7,8,9,10,11,12},
        legend pos=north west,
        grid=both,
    ]
    
    % ELECTRA
    \nextgroupplot[]
    \addplot[
        ybar, fill=cyan,
        ]
        coordinates {
            (0, 0.16961944048474759)(1, 0.17277126394964018)(2, 0.2230462603133715)(3, 0.26742274563603546)(4, 0.30471455774649664)(5, 0.3852178048848588)(6, 0.4243945608620198)(7, 0.4775646295364916)(8, 0.5350450176539249)(9, 0.5301312348234369)(10, 0.5491920151313351)(11, 0.6212800673903565)(12, 0.6130691845013366)
        };
    \addplot[
        ybar, pattern=north east lines, pattern color=cyan
        ]
        coordinates {
            (0, -0.0013954445184193032)(1, 0.003782236651342058)(2, 0.009014275187109844)(3, -0.004826877075735203)(4, 0.03662855729990311)(5, 0.05115382496650517)(6, 0.08847623385027586)(7, 0.07480924477721851)(8, 0.06279575603019594)(9, 0.10482068344124806)(10, 0.10912836311723662)(11, 0.07022538120872845)(12, 0.09136692241642586)
        };
    \legend{ELECTRA, Fine-Tuned ELECTRA}
    
    % BERT
    \nextgroupplot[]
    \addplot[
        ybar, fill=blue,
        ]
        coordinates {
            (0, 0.1556914671770591)(1, 0.1537150236323076)(2, 0.17728074365969818)(3, 0.2479918121594101)(4, 0.32136760931221914)(5, 0.3864685692400811)(6, 0.3911232867925886)(7, 0.43702767073761534)(8, 0.4654658584763908)(9, 0.47428224772244454)(10, 0.4680511526127468)(11, 0.47155524547331656)(12, 0.4130579202563726)
        };
    \addplot[
        ybar, pattern=north east lines, pattern color=blue
        ]
        coordinates {
            (0, -0.004600830438181797)(1, 0.012272949807237865)(2, 0.007980576805299705)(3, 0.022134712354262592)(4, 0.03693753746947859)(5, 0.03085505685469969)(6, 0.09654798042044793)(7, 0.09789033912363815)(8, 0.14261766811237048)(9, 0.11229806797327879)(10, 0.12510602043179703)(11, 0.12577659826015214)(12, 0.1847828534277482)
        };
    \legend{BERT, Fine-Tuned BERT}
    
    % XLNet
    \nextgroupplot[]
    \addplot[
        ybar, fill=magenta,
        ]
        coordinates {
            (0, 0.15124849710977026)(1, 0.12092178075845525)(2, 0.21595206514773704)(3, 0.23425864120888173)(4, 0.27545659346941376)(5, 0.3225892878730812)(6, 0.3573161278655267)(7, 0.34254883842611045)(8, 0.3218676439365221)(9, 0.27248770819981843)(10, 0.2711945584758675)(11, 0.288200849020865)(12, 0.28091437004681186)
        };
    \addplot[
        ybar, pattern=north east lines, pattern color=magenta
        ]
        coordinates {
            (0, -0.013040454234795856)(1, 0.04058536200521323)(2, 0.006091755031055357)(3, 0.10072495974998308)(4, 0.16045784604104607)(5, 0.17272336922475484)(6, 0.1611469276681996)(7, 0.1870486100242536)(8, 0.21083732024864899)(9, 0.251376718354748)(10, 0.2568249834294075)(11, 0.23397524052703672)(12, 0.23979146746771363)
        };
    \legend{XLNet, Fine-Tuned XLNet}
        
    \end{groupplot}
    \end{tikzpicture}
    
    \caption{Layer-wise comparison of the performance scores on CoLA dataset across the representations of the pre-trained and fine-tuned models.}
    \label{fig:feature_cola}
\end{figure}
\begin{figure}[ht!]
\centering
    \begin{tikzpicture}
    \begin{axis}[
        width=0.47\textwidth,
        height=8cm,
        xlabel={Layer},
        ylabel={RSA Similarity},
        title style={font=\small},
        label style={font=\small},
        legend style={font=\small},
        % xmin=0, %xmax=100,
        ymin=-0.07, ymax=1.03,
        xtick={0,2,4,6,8,10,12},
        legend pos=south west,
        grid=both,
    ]
    
    % BERT
    \addplot[
        color=blue,
        mark=square*,
        thick,
        error bars/.cd, y dir=both, y explicit,
        ]
        coordinates {
        (0, 0.9999059902297126) +=(0, 1.53912438286552e-05) -=(0, 2.3351775275348707e-05)
(1, 0.9975461628701952) +=(1, 0.000696811411115883) -=(1, 0.0011073615815904159)
(2, 0.9928611318270365) +=(2, 0.0028410752614339563) -=(2, 0.004236141840616825)
(3, 0.9912224743101332) +=(3, 0.001711633470323326) -=(3, 0.003158960077497719)
(4, 0.9912762443224589) +=(4, 0.0007942517598470422) -=(4, 0.0011808077494303015)
(5, 0.9889982276492648) +=(5, 0.00048691696590852374) -=(5, 0.000815682941012863)
(6, 0.9875318540467156) +=(6, 0.0007374750243293038) -=(6, 0.001166774166954876)
(7, 0.9866549372673035) +=(7, 0.0010374188423156738) -=(7, 0.0013816356658935547)
(8, 0.9831028580665588) +=(8, 0.0008909106254577637) -=(8, 0.0012874007225036621)
(9, 0.9555608232816061) +=(9, 0.002971986929575565) -=(9, 0.003571649392445919)
(10, 0.9248775111304389) +=(10, 0.009462667836083294) -=(10, 0.009763167964087605)
(11, 0.817523353629642) +=(11, 0.04921496576733064) -=(11, 0.04388671451144743)
(12, 0.3215470181571113) +=(12, 0.061920298470391155) -=(12, 0.047357188330756306)
        };

    % XLNet
    \addplot[
        color=magenta,
        mark=*,
        thick,
        error bars/.cd, y dir=both, y explicit,
        ]
        coordinates {
        (0, 0.9999592105547587) +=(0, 9.079774220821513e-06) -=(0, 6.41743342078005e-06)
(1, 0.9137102961540222) +=(1, 0.018102824687957764) -=(1, 0.015135645866394043)
(2, 0.89399121205012) +=(2, 0.014322121938069698) -=(2, 0.01053265730539954)
(3, 0.8532754646407233) +=(3, 0.004702581299675823) -=(3, 0.004837976561652302)
(4, 0.910895950264401) +=(4, 0.019350462489657883) -=(4, 0.021192140049404617)
(5, 0.8510020971298218) +=(5, 0.012782394886016846) -=(5, 0.016744017601013184)
(6, 0.8148995571666293) +=(6, 0.013841695255703446) -=(6, 0.0077897840076022185)
(7, 0.7921347154511346) +=(7, 0.023070798979865192) -=(7, 0.019494427575005413)
(8, 0.7118648886680603) +=(8, 0.03661984205245972) -=(8, 0.020865023136138916)
(9, 0.6885196301672194) +=(9, 0.0274036195543077) -=(9, 0.03014126751157975)
(10, 0.6176232894261678) +=(10, 0.03744951883951819) -=(10, 0.03966279824574792)
(11, 0.4720376663737827) +=(11, 0.050650364822811544) -=(11, 0.05983739429050022)
(12, 0.16056417922178903) +=(12, 0.024913509686787932) -=(12, 0.0313634971777598)
        };
    
    % ELECTRA
    \addplot[color=cyan, mark=triangle*, thick, error bars/.cd, y dir=both, y explicit,]
        coordinates {
        (0, 0.999951011604733) +=(0, 7.861190371993843e-06) -=(0, 1.234478420686358e-05)
(1, 0.9973156054814657) +=(1, 0.0009015599886575965) -=(1, 0.000852962334950802)
(2, 0.9926949209637113) +=(2, 0.0005084209971957243) -=(2, 0.0006599492496914827)
(3, 0.9915565384758843) +=(3, 0.001699255572425007) -=(3, 0.0013932718171013603)
(4, 0.9916935695542229) +=(4, 0.0019656287299262276) -=(4, 0.001557423008812786)
(5, 0.9881923927201165) +=(5, 0.002536700831519245) -=(5, 0.0024091137780083427)
(6, 0.9912769993146261) +=(6, 0.0008809963862100867) -=(6, 0.0011233290036519739)
(7, 0.9892157117525736) +=(7, 0.0016042788823446008) -=(7, 0.0022768775622049597)
(8, 0.9853278994560242) +=(8, 0.003672778606414795) -=(8, 0.0035878419876098633)
(9, 0.98044177558687) +=(9, 0.004129085275861977) -=(9, 0.0030865338113572616)
(10, 0.9777941836251153) +=(10, 0.004886196719275593) -=(10, 0.0048565997017754325)
(11, 0.9345603651470609) +=(11, 0.028159313731723312) -=(11, 0.03960097498363924)
(12, 0.387056161959966) +=(12, 0.11103172103563946) -=(12, 0.11049719651540119)
        };
    
    % ERROR Regions
    % BERT
    \addplot [name path=upper_bert, fill=none, draw=none]  
    coordinates {
        (0, 0.9999213814735413)(1, 0.998242974281311)(2, 0.9957022070884705)(3, 0.9929341077804565)(4, 0.9920704960823059)(5, 0.9894851446151733)(6, 0.9882693290710449)(7, 0.9876923561096191)(8, 0.9839937686920166)(9, 0.9585328102111816)(10, 0.9343401789665222)(11, 0.8667383193969727)(12, 0.38346731662750244)
    };
    \addplot [name path=lower_bert, fill=none, draw=none]  
    coordinates {
        (0, 0.9998826384544373)(1, 0.9964388012886047)(2, 0.9886249899864197)(3, 0.9880635142326355)(4, 0.9900954365730286)(5, 0.988182544708252)(6, 0.9863650798797607)(7, 0.9852733016014099)(8, 0.9818154573440552)(9, 0.9519891738891602)(10, 0.9151143431663513)(11, 0.7736366391181946)(12, 0.274189829826355)
    };
    \addplot[blue!40] fill between[of=lower_bert and upper_bert];
    
    % XLNet
    \addplot [name path=upper_xlnet, fill=none, draw=none]  
    coordinates {
        (0, 0.9999682903289795)(1, 0.93181312084198)(2, 0.9083133339881897)(3, 0.8579780459403992)(4, 0.9302464127540588)(5, 0.8637844920158386)(6, 0.8287412524223328)(7, 0.8152055144309998)(8, 0.74848473072052)(9, 0.7159232497215271)(10, 0.655072808265686)(11, 0.5226880311965942)(12, 0.18547768890857697)
    };
    \addplot [name path=lower_xlnet, fill=none, draw=none]  
    coordinates {
        (0, 0.9999527931213379)(1, 0.8985746502876282)(2, 0.8834585547447205)(3, 0.848437488079071)(4, 0.8897038102149963)(5, 0.8342580795288086)(6, 0.8071097731590271)(7, 0.7726402878761292)(8, 0.6909998655319214)(9, 0.6583783626556396)(10, 0.5779604911804199)(11, 0.41220027208328247)(12, 0.12920068204402924)
    };
    \addplot[magenta!40] fill between[of=lower_xlnet and upper_xlnet];
    
    % ELECTRA
    \addplot [name path=upper_electra, fill=none, draw=none]  
    coordinates {
        (0, 0.999958872795105)(1, 0.9982171654701233)(2, 0.993203341960907)(3, 0.9932557940483093)(4, 0.9936591982841492)(5, 0.9907290935516357)(6, 0.9921579957008362)(7, 0.9908199906349182)(8, 0.989000678062439)(9, 0.9845708608627319)(10, 0.9826803803443909)(11, 0.9627196788787842)(12, 0.49808788299560547)
    };
    \addplot [name path=lower_electra, fill=none, draw=none]  
    coordinates {
        (0, 0.9999386668205261)(1, 0.9964626431465149)(2, 0.9920349717140198)(3, 0.990163266658783)(4, 0.9901361465454102)(5, 0.9857832789421082)(6, 0.9901536703109741)(7, 0.9869388341903687)(8, 0.9817400574684143)(9, 0.9773552417755127)(10, 0.9729375839233398)(11, 0.8949593901634216)(12, 0.2765589654445648)
    };
    \addplot[cyan!40] fill between[of=lower_electra and upper_electra];
    
    \legend{BERT, XLNET, ELECTRA}
        
    \end{axis}
    \end{tikzpicture}
    
    \caption{Comparison of the representations of BERT, XLNet, and ELECTRA base, with their respective fine-tuned model on CoLA dataset.}
    \label{fig:rsa_cola}
\end{figure}

% \newpage

\begin{table*}[b]
\centering
\begin{tabular}{lccc}
\toprule
Task & Labels & Number of targets \\
\midrule
Dependency Labeling & 49 & 203919 / 25110 / 25049  \\
Named Entity Recognition & 18 & 128738 / 20354 / 12586 \\
Semantic Role Labeling & 66 & 598983 / 83362 / 61716 \\
Coreference Resolution & 2 & 207830 / 26333 / 27800 \\
Relation Classification & 19 & 6851 / 1149 / 2717 \\
\bottomrule
\end{tabular}
\vspace{-1ex}
\caption{Dataset statistics for all five core tasks used in probing. Numbers of targets are given for train / dev / test sets.}
\label{tab:dataset_stat}
\end{table*}
\newpage
\subsection{Results for SST-2 Dataset}
Figures \ref{fig:feature_sst2} and \ref{fig:rsa_sst2} report the performance and RSA results before and after fine-tuning on SST-2 dataset.
\begin{figure}[h]
\centering
    \begin{tikzpicture}
    \begin{groupplot}[
        group style={
            group name=my plots,
            group size=1 by 3,
            xlabels at=edge bottom,
            xticklabels at=edge bottom,
            vertical sep=0pt,
        },
        ybar stacked, 
        width=0.49\textwidth,
        height=5cm,
        xlabel={Layer},
        ylabel={Accuracy},
        title style={font=\small},
        label style={font=\small},
        legend style={font=\tiny},
        % xmin=0, %xmax=100,
        ymin=0.72, ymax=0.97,
        xtick={0,...,12},
        % xticklabel interval boundaries,
        % ytick={0,1,2,3,4,5,6,7,8,9,10,11,12},
        legend pos=north west,
        grid=both,
    ]
    
    % ELECTRA
    \nextgroupplot[]
    \addplot[
        ybar, fill=cyan,
        ]
        coordinates {
            (0, 0.8348623853211009)(1, 0.8405963302752294)(2, 0.8337155963302753)(3, 0.8451834862385321)(4, 0.856651376146789)(5, 0.8704128440366973)(6, 0.875)(7, 0.9105504587155964)(8, 0.9185779816513762)(9, 0.930045871559633)(10, 0.9311926605504587)(11, 0.9277522935779816)(12, 0.9059633027522935)
        };
    \addplot[
        ybar, pattern=north east lines, pattern color=cyan
        ]
        coordinates {
            (0, 0.0011467889908256534)(1, 0.0011467889908256534)(2, 0.016055045871559592)(3, 0.012614678899082521)(4, -0.006880733944954143)(5, 0.004587155963302725)(6, 0.002293577981651418)(7, -0.002293577981651418)(8, 0.011467889908256867)(9, 0.016055045871559592)(10, 0.020642201834862428)(11, 0.025229357798165153)(12, 0.048165137614679)
        };
    \legend{ELECTRA, Fine-Tuned ELECTRA}
    
    % BERT
    \nextgroupplot[]
    \addplot[
        ybar, fill=blue,
        ]
        coordinates {
            (0, 0.8165137614678899)(1, 0.8222477064220184)(2, 0.8142201834862385)(3, 0.819954128440367)(4, 0.8337155963302753)(5, 0.8463302752293578)(6, 0.8360091743119266)(7, 0.8520642201834863)(8, 0.8646788990825688)(9, 0.8807339449541285)(10, 0.8887614678899083)(11, 0.8841743119266054)(12, 0.8795871559633027)
        };
    \addplot[
        ybar, pattern=north east lines, pattern color=blue
        ]
        coordinates {
            (0, 0.008027522935779796)(1, 0.00917431192660545)(2, 0.010321100917431214)(3, 0.002293577981651418)(4, -0.0034403669724770714)(5, 0.00917431192660556)(6, 0.032110091743119296)(7, 0.04013761467889909)(8, 0.028669724770642224)(9, 0.03096330275229353)(10, 0.037844036697247674)(11, 0.04587155963302758)(12, 0.05160550458715596)
        };
    \legend{BERT, Fine-Tuned BERT}
    
    % XLNet
    \nextgroupplot[]
    \addplot[
        ybar, fill=magenta,
        ]
        coordinates {
            (0, 0.8188073394495413)(1, 0.8405963302752294)(2, 0.8451834862385321)(3, 0.8600917431192661)(4, 0.8600917431192661)(5, 0.8841743119266054)(6, 0.8967889908256881)(7, 0.9013761467889908)(8, 0.9071100917431193)(9, 0.8979357798165137)(10, 0.8979357798165137)(11, 0.8807339449541285)(12, 0.8623853211009175)
        };
    \addplot[
        ybar, pattern=north east lines, pattern color=magenta
        ]
        coordinates {
            (0, 0.011467889908256867)(1, 0.0011467889908256534)(2, 0.0)(3, 0.02178899082568808)(4, 0.04013761467889909)(5, 0.020642201834862428)(6, 0.02178899082568808)(7, 0.014908256880733939)(8, 0.025229357798165153)(9, 0.044724770642201817)(10, 0.044724770642201817)(11, 0.059633027522935755)(12, 0.07568807339449535)
        };
    \legend{XLNet, Fine-Tuned XLNet}
        
    \end{groupplot}
    \end{tikzpicture}
    
    \caption{Layer-wise comparison of the performance scores on SST-2 dataset across the representations of the pre-trained and fine-tuned models.}
    \label{fig:feature_sst2}
\end{figure}
\begin{figure}[ht!]
\centering
    \begin{tikzpicture}
    \begin{axis}[
        width=0.47\textwidth,
        height=8cm,
        xlabel={Layer},
        ylabel={RSA Similarity},
        title style={font=\small},
        label style={font=\small},
        legend style={font=\small},
        % xmin=0, %xmax=100,
        ymin=-0.07, ymax=1.03,
        xtick={0,2,4,6,8,10,12},
        legend pos=south west,
        grid=both,
    ]
    
    % BERT
    \addplot[
        color=blue,
        mark=square*,
        thick,
        error bars/.cd, y dir=both, y explicit,
        ]
        coordinates {
        (0, 0.9995349910524156) +=(0, 0.0001861188146803361) -=(0, 0.0001554157998826522)
(1, 0.9924371242523193) +=(1, 0.0004907846450805664) -=(1, 0.00048726797103881836)
(2, 0.9806196490923563) +=(2, 0.001468499501546261) -=(2, 0.0017236868540445593)
(3, 0.9800252980656095) +=(3, 0.00339549117618132) -=(3, 0.002542442745632645)
(4, 0.9817675484551324) +=(4, 0.00472395287619698) -=(4, 0.0037316812409294853)
(5, 0.9777664542198181) +=(5, 0.0033537745475769043) -=(5, 0.003762364387512207)
(6, 0.976051496134864) +=(6, 0.0020436777008904228) -=(6, 0.002534674273596882)
(7, 0.9750986496607462) +=(7, 0.004125972588857052) -=(7, 0.002886931101481083)
(8, 0.9633281893200345) +=(8, 0.007140152984195236) -=(8, 0.005548304981655594)
(9, 0.9190432826677958) +=(9, 0.017784496148427364) -=(9, 0.014789799849192264)
(10, 0.7899832328160604) +=(10, 0.05107734600702918) -=(10, 0.04414590199788415)
(11, 0.5963386230998569) +=(11, 0.08512634701199007) -=(11, 0.06639265351825285)
(12, 0.18295002977053323) +=(12, 0.04898597796758017) -=(12, 0.043244938055674226)
        };

    % XLNet
    \addplot[
        color=magenta,
        mark=*,
        thick,
        error bars/.cd, y dir=both, y explicit,
        ]
        coordinates {
        (0, 0.9998420543140836) +=(0, 0.00011222892337370283) -=(0, 6.938642925691241e-05)
(1, 0.7864458627170987) +=(1, 0.10970842176013518) -=(1, 0.07398778862423372)
(2, 0.8337020940250821) +=(2, 0.06635301642947722) -=(2, 0.04508156246609163)
(3, 0.8056948251194425) +=(3, 0.08151131206088591) -=(3, 0.05752719110912752)
(4, 0.8448745873239305) +=(4, 0.034346017572614906) -=(4, 0.01932188537385726)
(5, 0.7882307039366828) +=(5, 0.03290296925438774) -=(5, 0.02572934495078194)
(6, 0.708005686601003) +=(6, 0.024612704912821415) -=(6, 0.03142501910527551)
(7, 0.6868339776992798) +=(7, 0.03272479772567749) -=(7, 0.02577906847000122)
(8, 0.5459349950154623) +=(8, 0.057896653811136845) -=(8, 0.03974231084187829)
(9, 0.40045201116138035) +=(9, 0.14995140499538845) -=(9, 0.10562463932567173)
(10, 0.2851122319698334) +=(10, 0.13962647318840027) -=(10, 0.09953099489212036)
(11, 0.16027720934814876) +=(11, 0.17126569979720646) -=(11, 0.11089758119649357)
(12, 0.08479384167326821) +=(12, 0.019805522428618536) -=(12, 0.010058423711193934)
        };
    
    % ELECTRA
    \addplot[color=cyan, mark=triangle*, thick, error bars/.cd, y dir=both, y explicit,]
        coordinates {
        (0, 0.9995764957533942) +=(0, 4.1855706108928636e-05) -=(0, 4.5941935645221754e-05)
(1, 0.9900649521085951) +=(1, 0.0012151267793443221) -=(1, 0.001118097040388344)
(2, 0.9809434215227762) +=(2, 0.002146442731221554) -=(2, 0.004314581553141239)
(3, 0.9791028698285421) +=(3, 0.001638809839884403) -=(3, 0.002739270528157589)
(4, 0.9805654022428725) +=(4, 0.0026672151353623885) -=(4, 0.0030460092756483537)
(5, 0.9733633266554939) +=(5, 0.003107739819420696) -=(5, 0.003391193019019245)
(6, 0.9763866133160062) +=(6, 0.0044505529933505095) -=(6, 0.004524177975124832)
(7, 0.965377926826477) +=(7, 0.0044899582862854) -=(7, 0.0055207014083862305)
(8, 0.9479454027281867) +=(8, 0.008016301525963665) -=(8, 0.006716953383551716)
(9, 0.9062405692206489) +=(9, 0.02168722285164726) -=(9, 0.0357525216208564)
(10, 0.6335204409228431) +=(10, 0.08416019214524162) -=(10, 0.15438656674491036)
(11, 0.2536253333091736) +=(11, 0.09291550517082214) -=(11, 0.12355612218379974)
(12, -0.015337628622849783) +=(12, 0.07702858870228131) -=(12, 0.048598902920881905)
        };
    
    % ERROR Regions
    % BERT
    \addplot [name path=upper_bert, fill=none, draw=none]  
    coordinates {
        (0, 0.999721109867096)(1, 0.9929279088973999)(2, 0.9820881485939026)(3, 0.9834207892417908)(4, 0.9864915013313293)(5, 0.981120228767395)(6, 0.9780951738357544)(7, 0.9792246222496033)(8, 0.9704683423042297)(9, 0.9368277788162231)(10, 0.8410605788230896)(11, 0.6814649701118469)(12, 0.2319360077381134)
    };
    \addplot [name path=lower_bert, fill=none, draw=none]  
    coordinates {
        (0, 0.999379575252533)(1, 0.9919498562812805)(2, 0.9788959622383118)(3, 0.9774828553199768)(4, 0.9780358672142029)(5, 0.9740040898323059)(6, 0.9735168218612671)(7, 0.9722117185592651)(8, 0.9577798843383789)(9, 0.9042534828186035)(10, 0.7458373308181763)(11, 0.529945969581604)(12, 0.139705091714859)
    };
    \addplot[blue!40] fill between[of=lower_bert and upper_bert];
    
    % XLNet
    \addplot [name path=upper_xlnet, fill=none, draw=none]  
    coordinates {
        (0, 0.9999542832374573)(1, 0.8961542844772339)(2, 0.9000551104545593)(3, 0.8872061371803284)(4, 0.8792206048965454)(5, 0.8211336731910706)(6, 0.7326183915138245)(7, 0.7195587754249573)(8, 0.6038316488265991)(9, 0.5504034161567688)(10, 0.42473870515823364)(11, 0.3315429091453552)(12, 0.10459936410188675)
    };
    \addplot [name path=lower_xlnet, fill=none, draw=none]  
    coordinates {
        (0, 0.9997726678848267)(1, 0.712458074092865)(2, 0.7886205315589905)(3, 0.7481676340103149)(4, 0.8255527019500732)(5, 0.7625013589859009)(6, 0.6765806674957275)(7, 0.6610549092292786)(8, 0.506192684173584)(9, 0.2948273718357086)(10, 0.185581237077713)(11, 0.0493796281516552)(12, 0.07473541796207428)
    };
    \addplot[magenta!40] fill between[of=lower_xlnet and upper_xlnet];
    
    % ELECTRA
    \addplot [name path=upper_electra, fill=none, draw=none]  
    coordinates {
        (0, 0.9996183514595032)(1, 0.9912800788879395)(2, 0.9830898642539978)(3, 0.9807416796684265)(4, 0.9832326173782349)(5, 0.9764710664749146)(6, 0.9808371663093567)(7, 0.9698678851127625)(8, 0.9559617042541504)(9, 0.9279277920722961)(10, 0.7176806330680847)(11, 0.3465408384799957)(12, 0.061690960079431534)
    };
    \addplot [name path=lower_electra, fill=none, draw=none]  
    coordinates {
        (0, 0.999530553817749)(1, 0.9889468550682068)(2, 0.976628839969635)(3, 0.9763635993003845)(4, 0.9775193929672241)(5, 0.9699721336364746)(6, 0.9718624353408813)(7, 0.9598572254180908)(8, 0.941228449344635)(9, 0.8704880475997925)(10, 0.47913387417793274)(11, 0.13006921112537384)(12, -0.06393653154373169)
    };
    \addplot[cyan!40] fill between[of=lower_electra and upper_electra];
    
    \legend{BERT, XLNET, ELECTRA}
        
    \end{axis}
    \end{tikzpicture}
    
    \caption{Comparison of the representations of BERT, XLNet, and ELECTRA base, with their respective fine-tuned model on SST-2 dataset.}
    \label{fig:rsa_sst2}
\end{figure}

\newpage
\subsection{Datasets Statistics}
The number of labels and targets of the NLP linguistic tasks we used in probing are reported in table \ref{tab:dataset_stat}.

\newpage
\begin{table*}[h!t!]
\small
\begin{center}
\begin{tabular}{c | c c | c c | c c} 
 \toprule
    % BERT & XLNet & ELECTRA \\
      & \multicolumn{2}{c|}{\textbf{BERT}} & \multicolumn{2}{c|}{\textbf{XLNet}} & \multicolumn{2}{c}{\textbf{ELECTRA}} \\
     Tasks & pre-trained & fine-tuned & pre-trained & fine-tuned & pre-trained & fine-tuned \\
    % \textbf{Task} & F1 Score & Compression & F1 Score & Compression & F1 Score & Compression \\
    \midrule
    \multirow{ 13}{*}{Dependencies} &  5.99 (186.7)  & 6.04 (185.1)  & 5.01 (223.1)  & 5.03 (222.3)  & 6.35 (176.0)  & 6.33 (176.5)  \\
    &  8.39 (133.3)  & 7.89 (141.7)  & 9.39 (119.1)  & 8.41 (132.9)  & 9.33 (119.8)  & 9.26 (120.8)  \\
    &  9.49 (117.8)  & 8.92 (125.3)  & 11.37 (98.3)  & 10.28 (108.8) & 10.71 (104.4) & 10.70 (104.5) \\
    &  10.75 (104.0) & 10.29 (108.7) & 13.18 (84.8)  & 12.44 (89.8)  & 10.97 (101.9) & 10.93 (102.3) \\
    &  12.75 (87.7)  & 12.12 (92.3)  & 13.98 (80.0)  & 13.14 (85.1)  & 11.75 (95.2)  & 11.55 (96.8)  \\
    &  13.57 (82.4)  & 12.88 (86.8)  & 14.00 (79.9)  & 13.41 (83.4)  & 13.54 (82.5)  & 13.57 (82.4)  \\
    &  14.45 (77.4)  & 13.45 (83.1)  & 14.13 (79.1)  & 13.54 (82.6)  & 14.42 (77.5)  & 14.36 (77.9)  \\
    &  14.95 (74.8)  & 13.81 (81.0)  & 13.57 (82.4)  & 13.11 (85.3)  & 15.65 (71.5)  & 15.04 (74.3)  \\
    &  15.25 (73.3)  & 13.61 (82.2)  & 13.30 (84.1)  & 12.19 (91.7)  & 16.15 (69.2)  & 15.15 (73.8)  \\
    &  14.34 (78.0)  & 12.46 (89.7)  & 12.72 (87.9)  & 10.93 (102.3) & 15.67 (71.3)  & 14.27 (78.3)  \\
    &  13.17 (84.9)  & 11.60 (96.4)  & 11.63 (96.2)  & 9.32 (120.0)  & 15.76 (70.9)  & 14.03 (79.7)  \\
    &  12.08 (92.5)  & 10.83 (103.3) & 10.42 (107.3) & 6.77 (165.1)  & 15.44 (72.4)  & 13.89 (80.5)  \\
    &  11.06 (101.1) & 9.76 (114.6)  & 7.26 (154.1)  & 3.08 (362.5)  & 13.62 (82.1)  & 11.98 (93.4)  \\
    \midrule
    \multirow{ 13}{*}{Entities} &  8.37 (62.6)  & 8.28 (63.3)  & 9.33 (56.2)  & 9.41 (55.7)  & 8.72 (60.1)  & 8.63 (60.7)  \\
    &  11.37 (46.1) & 10.86 (48.3) & 12.14 (43.2) & 11.67 (44.9) & 11.74 (44.6) & 11.47 (45.7) \\
    &  12.19 (43.0) & 11.77 (44.5) & 13.56 (38.7) & 13.57 (38.6) & 13.55 (38.7) & 13.24 (39.6) \\
    &  12.95 (40.5) & 12.48 (42.0) & 14.97 (35.0) & 14.92 (35.1) & 14.79 (35.4) & 14.30 (36.7) \\
    &  14.15 (37.0) & 13.65 (38.4) & 15.46 (33.9) & 15.79 (33.2) & 15.25 (34.4) & 15.02 (34.9) \\
    &  15.28 (34.3) & 14.76 (35.5) & 15.30 (34.3) & 15.91 (33.0) & 16.77 (31.3) & 15.99 (32.8) \\
    &  15.50 (33.8) & 14.92 (35.1) & 14.85 (35.3) & 15.63 (33.5) & 16.88 (31.1) & 15.79 (33.2) \\
    &  16.29 (32.2) & 15.32 (34.2) & 14.29 (36.7) & 15.61 (33.6) & 16.60 (31.6) & 15.63 (33.5) \\
    &  16.77 (31.3) & 15.58 (33.7) & 14.20 (36.9) & 15.13 (34.6) & 16.76 (31.3) & 15.44 (34.0) \\
    &  16.87 (31.1) & 15.33 (34.2) & 13.72 (38.2) & 14.09 (37.2) & 16.46 (31.9) & 14.83 (35.3) \\
    &  16.59 (31.6) & 14.72 (35.6) & 13.36 (39.2) & 12.92 (40.6) & 16.29 (32.2) & 14.22 (36.9) \\
    &  16.06 (32.7) & 14.55 (36.0) & 12.77 (41.0) & 10.46 (50.1) & 15.47 (33.9) & 13.59 (38.6) \\
    &  15.64 (33.5) & 13.60 (38.5) & 10.84 (48.4) & 5.38 (97.4)  & 14.10 (37.2) & 11.21 (46.8) \\
    \midrule
    \multirow{ 13}{*}{SRL} &  7.93 (445.7)  & 7.87 (449.3)  & 6.91 (511.4)  & 6.93 (510.3)  & 8.07 (438.4)  & 8.03 (440.1)  \\
    &  9.67 (365.7)  & 9.44 (374.7)  & 10.70 (330.5) & 10.33 (342.2) & 10.55 (335.0) & 10.48 (337.4) \\
    &  10.42 (339.2) & 10.26 (344.6) & 11.90 (297.1) & 11.68 (302.6) & 11.64 (303.7) & 11.51 (307.3) \\
    &  11.38 (310.8) & 11.27 (313.8) & 12.95 (273.0) & 12.98 (272.5) & 12.07 (292.9) & 11.94 (296.0) \\
    &  12.45 (284.0) & 12.40 (285.2) & 13.32 (265.5) & 13.39 (264.0) & 12.77 (276.8) & 12.59 (280.8) \\
    &  13.25 (266.8) & 12.91 (273.8) & 13.23 (267.3) & 13.36 (264.6) & 13.79 (256.4) & 13.55 (261.0) \\
    &  13.73 (257.6) & 13.26 (266.7) & 13.05 (270.9) & 13.25 (266.8) & 14.13 (250.2) & 13.91 (254.2) \\
    &  13.94 (253.7) & 13.30 (265.9) & 12.50 (282.8) & 12.66 (279.3) & 14.44 (244.8) & 14.03 (252.1) \\
    &  13.88 (254.8) & 13.06 (270.7) & 12.05 (293.4) & 12.13 (291.4) & 14.32 (246.8) & 13.59 (260.1) \\
    &  13.27 (266.5) & 12.35 (286.2) & 11.57 (305.7) & 11.30 (312.8) & 13.94 (253.6) & 13.09 (270.1) \\
    &  12.54 (281.8) & 11.64 (303.7) & 11.06 (319.8) & 10.02 (352.7) & 13.78 (256.6) & 12.79 (276.4) \\
    &  12.00 (294.6) & 11.23 (314.8) & 10.36 (341.2) & 8.16 (433.3)  & 13.52 (261.4) & 12.40 (285.2) \\
    &  11.46 (308.5) & 10.58 (334.3) & 8.40 (420.9)  & 4.60 (769.0)  & 12.47 (283.5) & 11.32 (312.3) \\
    \midrule
    \multirow{ 13}{*}{Coreference} &  2.74 (74.2) & 2.75 (73.8) & 2.65 (76.6) & 2.63 (77.3)  & 2.73 (74.4) & 2.75 (73.8) \\
    &  2.96 (68.5) & 2.94 (69.0) & 3.03 (67.0) & 2.91 (69.9)  & 3.06 (66.4) & 3.06 (66.2) \\
    &  3.11 (65.2) & 3.06 (66.4) & 3.31 (61.3) & 3.24 (62.6)  & 3.23 (62.9) & 3.22 (63.0) \\
    &  3.29 (61.6) & 3.31 (61.3) & 3.63 (55.9) & 3.63 (56.0)  & 3.33 (60.9) & 3.35 (60.5) \\
    &  3.59 (56.5) & 3.57 (56.8) & 3.88 (52.3) & 3.87 (52.4)  & 3.57 (56.8) & 3.57 (56.9) \\
    &  3.76 (54.0) & 3.76 (54.0) & 3.92 (51.8) & 4.00 (50.7)  & 4.04 (50.2) & 3.93 (51.6) \\
    &  3.95 (51.4) & 3.83 (53.0) & 3.95 (51.4) & 4.03 (50.3)  & 4.29 (47.3) & 4.22 (48.0) \\
    &  4.14 (49.1) & 3.98 (51.0) & 3.94 (51.5) & 4.07 (49.9)  & 4.56 (44.5) & 4.43 (45.8) \\
    &  4.43 (45.8) & 4.18 (48.5) & 3.93 (51.6) & 4.04 (50.3)  & 5.06 (40.1) & 4.78 (42.4) \\
    &  4.58 (44.3) & 4.13 (49.2) & 3.97 (51.1) & 3.95 (51.3)  & 5.44 (37.3) & 4.93 (41.2) \\
    &  4.49 (45.2) & 4.00 (50.7) & 3.78 (53.7) & 3.54 (57.3)  & 5.88 (34.5) & 5.09 (39.9) \\
    &  4.32 (47.0) & 3.85 (52.7) & 3.50 (58.0) & 3.11 (65.2)  & 5.68 (35.7) & 4.95 (41.0) \\
    &  4.09 (49.6) & 3.67 (55.3) & 2.82 (72.1) & 2.01 (101.1) & 4.76 (42.7) & 4.26 (47.6) \\
    \midrule
    \multirow{ 13}{*}{Relations} &  1.51 (18.9) & 1.52 (18.7) & 1.66 (17.1) & 1.65 (17.2) & 1.63 (17.4) & 1.61 (17.6) \\
    &  1.80 (15.8) & 1.73 (16.4) & 1.95 (14.5) & 1.85 (15.4) & 1.85 (15.3) & 1.86 (15.3) \\
    &  1.91 (14.9) & 1.85 (15.4) & 2.16 (13.1) & 2.04 (13.9) & 2.00 (14.2) & 2.01 (14.1) \\
    &  2.12 (13.4) & 2.02 (14.1) & 2.51 (11.3) & 2.43 (11.7) & 2.18 (13.0) & 2.15 (13.2) \\
    &  2.31 (12.3) & 2.29 (12.4) & 2.59 (11.0) & 2.58 (11.0) & 2.34 (12.2) & 2.33 (12.2) \\
    &  2.48 (11.5) & 2.42 (11.8) & 2.68 (10.6) & 2.67 (10.7) & 2.80 (10.1) & 2.69 (10.6) \\
    &  2.51 (11.3) & 2.56 (11.1) & 2.89 (9.8)  & 2.91 (9.8)  & 2.89 (9.8)  & 2.83 (10.0) \\
    &  2.74 (10.4) & 2.76 (10.3) & 2.90 (9.8)  & 3.00 (9.5)  & 3.04 (9.4)  & 2.94 (9.7)  \\
    &  2.98 (9.5)  & 2.95 (9.6)  & 2.95 (9.6)  & 3.07 (9.3)  & 3.37 (8.4)  & 3.26 (8.7)  \\
    &  3.04 (9.4)  & 3.03 (9.4)  & 2.95 (9.6)  & 2.91 (9.8)  & 3.34 (8.5)  & 3.19 (8.9)  \\
    &  2.93 (9.7)  & 2.93 (9.7)  & 2.97 (9.6)  & 2.69 (10.6) & 3.32 (8.6)  & 3.14 (9.0)  \\
    &  2.77 (10.3) & 2.86 (9.9)  & 2.82 (10.1) & 2.36 (12.1) & 2.98 (9.5)  & 2.83 (10.0) \\
    &  2.57 (11.1) & 2.69 (10.6) & 2.33 (12.2) & 1.71 (16.6) & 2.57 (11.0) & 2.30 (12.3) \\
 \bottomrule
\end{tabular}
\end{center}
\captionsetup{aboveskip=0pt}
\caption{
Cross-model MDL compression in pre-trained and fine-tuned models on MNLI dataset. The corresponding codelengths are presented in the brackets. Layers are 0 to 12 from top to bottom.}

\label{tab:compressions_appendix}
\end{table*}

% \multirow{ 13}{*}{Dependencies}
% \multirow{ 13}{*}{Entities}
% \multirow{ 13}{*}{SRL}
% \multirow{ 13}{*}{Coreference}
% \multirow{ 13}{*}{Relations}

\end{document}